\pdfoutput=1

\documentclass[11pt]{article}
\usepackage[table,xcdraw]{xcolor}
\usepackage[]{acl}

\usepackage{times}
\usepackage{latexsym}
\usepackage{graphicx}
\usepackage{subcaption}
\usepackage{multirow}
\usepackage{array}
\usepackage{booktabs}
\usepackage{url}
\usepackage{tablefootnote}
\usepackage{relsize}
\usepackage{siunitx}
\usepackage{amsmath}
\usepackage{fancyhdr}

\usepackage[T1]{fontenc}

\usepackage[utf8]{inputenc}

\usepackage{microtype}
\usepackage[normalem]{ulem}
\useunder{\uline}{\ul}{}

\usepackage{inconsolata}
\newcommand{\dec}[1]{\textcolor{red}{(\smaller $\downarrow$#1\%)}}
\fancypagestyle{specialfooter}{
  \fancyhead{} 
  
  \fancyfoot[C]{To appear in  \textit{Proceedings of the Association for Computational Linguistics \textbf{(ACL 2024)}}}}
\title{A Ship of Theseus:\\Curious Cases of Paraphrasing in LLM-Generated Texts}

\author{
Nafis Irtiza Tripto\textsuperscript{1} 
\hspace{0.1in} 
Saranya Venkatraman\textsuperscript{1}
\hspace{0.1in} 
Dominik Macko\textsuperscript{2}
\hspace{0.1in} 
Robert Moro\textsuperscript{2}\\
\textbf{Ivan Srba}\textsuperscript{2}
\hspace{0.1in} 
\textbf{Adaku Uchendu}\textsuperscript{3}
\hspace{0.1in} 
\textbf{Thai Le}\textsuperscript{4}
\hspace{0.1in} 
\textbf{Dongwon Lee}\textsuperscript{1}\\    
        \textsuperscript{1}The Pennsylvania State University, USA\\ 
       \textsuperscript{1}\texttt {\{nit5154,saranyav,dongwon\}@psu.edu} \\
       \textsuperscript{2}Kempelen Institute of Intelligent Technologies, Slovakia \\ 
       \textsuperscript{2}\texttt {\{dominik.macko,robert.moro,ivan.srba\}@kinit.sk} \\
         \textsuperscript{3}MIT Lincoln Laboratory, USA , \textsuperscript{4}Indiana University, USA\\ 
          \textsuperscript{3}\texttt {adaku.uchendu@ll.mit.edu}, \textsuperscript{4}\texttt {tle@iu.edu}  \\ 
  }

\begin{document}
\thispagestyle{specialfooter}
\maketitle
\begin{abstract}
In the realm of text manipulation and linguistic transformation, the question of {\em authorship} has been a subject of fascination and philosophical inquiry. Much like the \textbf{Ship of Theseus paradox}, which ponders whether a ship remains the same when each of its original planks is replaced, our research delves into an intriguing question: \textit{Does a text retain its original authorship when it undergoes numerous paraphrasing iterations?} Specifically, since Large Language Models (LLMs) have demonstrated remarkable proficiency in both the generation of original content and the modification of human-authored texts, a pivotal question emerges concerning the determination of authorship in instances where LLMs or similar paraphrasing tools are employed to rephrase the text--i.e., \textit{whether authorship should be attributed to the original human author or the AI-powered tool.} Therefore, we embark on a philosophical voyage through the seas of language and authorship to unravel this intricate puzzle. Using a computational approach, we discover that the diminishing performance in text classification models, with each successive paraphrasing iteration, is closely associated with the extent of deviation from the original author's style, thus provoking a reconsideration of the current notion of authorship. 
\end{abstract}


\section{Introduction}
\label{sec-introduction}

\begin{figure}[t!]
    \centering
    \includegraphics[width=0.98\linewidth]{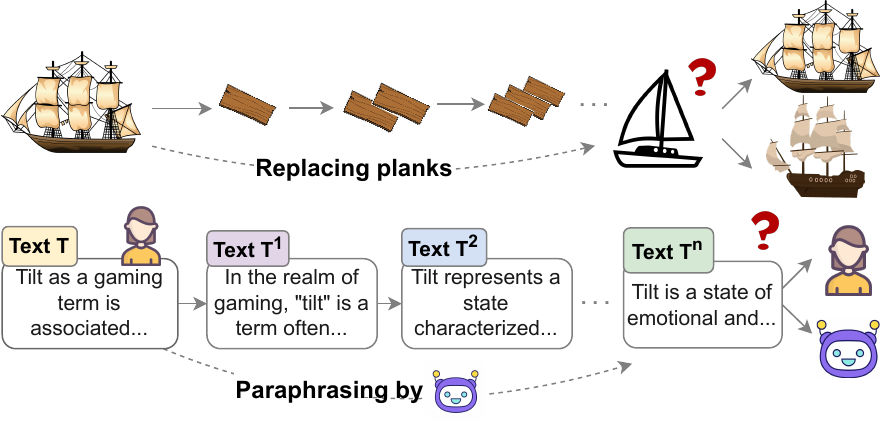}
    \caption{Ship of Theseus paradox in text paraphrasing scenario: who should be considered the author of $T^n$?}
    \label{fig_teaser_figure}
\end{figure}
The Ship of Theseus paradox 
is a philosophical thought experiment \citep{scaltsas1980ship_of_theseus} that questions the concept of originality and change over time. The paradox begins with the premise that a ship, called the \textbf{Ship of Theseus}, gradually has all its planks replaced over time with new, identical planks. The paradox then poses the question: \textit{Is the fully modified ship, with none of its original parts remaining, still the Ship of Theseus, or is it an entirely different ship?} Just like the Ship of Theseus, our study involves the successive transformation of text through paraphrasing as illustrated in Figure \ref{fig_teaser_figure}. Each paraphrase iteration can be seen as a replacement of linguistic ``planks.'' We aim to explore whether, like the Ship of Theseus, the essence of the original authorship remains intact or whether it morphs into something entirely new. 

\begin{figure*}[h]
    \centering
    \includegraphics[width=0.8\linewidth]{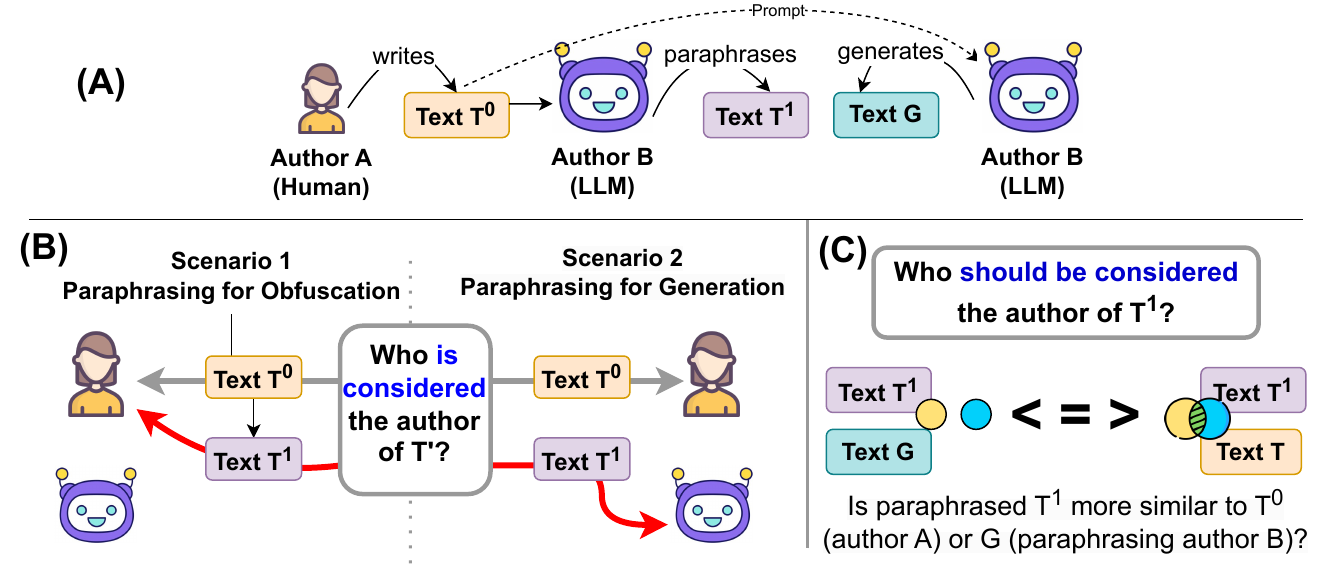}
    \caption{(A) indicates how LLMs can paraphrase as well as generate texts, (B) portrays the two alternative scenarios regarding authorship, and (C) shows how authorship should be determined.}
    \label{fig_motivation_figure}
\end{figure*}

Paraphrasing involves rewriting texts to convey the same meaning while employing different words or sentence structures \citep{bhagat2013paraphrase}. Although paraphrasing has long been employed to enhance writing, it has been the subject of ongoing ethical and plagiarism-related debates \citep{prentice2018paraphrasing,roe2022automated}. Nevertheless, paraphrasing has always been considered a tool to aid in rewriting content rather than generating entirely original material. However, recent advancements in LLMs have altered this paradigm as they can function as paraphrasers while also autonomously generating original content without explicit prompts. As illustrated in the examples in Figure \ref{fig_motivation_figure}, a situation will arise in contemporary times where paraphrasing a text ($T^0$) using an LLM to produce the paraphrased version ($T^1$) might closely resemble the text ($G$) independently generated by the LLM  on the same subject matter. Consequently, this situation prompts inquiries about the authorship of text $T^1$, akin to the philosophical dilemma posed by the Ship of Theseus.

Two contrasting perspectives on this matter are evident within the existing literature (Figure \ref{fig_motivation_figure}). Paraphrasing has often been employed as a text obfuscation or perturbation method \citep{potthast2016author,bevendorff2019heuristic,bevendorff2020divergence}. In line with this perspective, several studies \citep{krishna2024paraphrasing,sadasivan2023can,hu2023radar} argue that the weakness of a text classifier or AI-text detector is evident if it fails to attribute a paraphrased text to its original source precisely. Thus, these studies assume that \textbf{authorship remains the same after paraphrasing}. Conversely, paraphrasing can also serve as a text-generation technique. A growing number of recent studies \citep{yu2023cheat,zhang2024benchmarking,lucas2023fighting} utilize LLMs to rewrite human-generated texts through paraphrasing to create AI-generated datasets. Consequently, these studies presume that \textbf{authorship changes after paraphrasing}.

\textcolor{black}{
In this research study, we systematically investigate two distinct scenarios, as illustrated in Figure \ref{fig_motivation_figure}, and assess the impact of considering them as ground truth on the performance of authorship attribution, a multi-class classification task aimed at determining the author of a given text from a set of authors, and human vs. AI text detection, a binary classification task. The efficacy of text classification is inherently shaped by two pivotal factors: \textbf{content}, representing the subject matter or thematic focus of the text, and \textbf{style}, reflecting the unique manner of expression \citep{sari2018topic_or_style}. Therefore, we delve into the effects of successive paraphrasing on the original style and content of the text and its consequent implications for the performance of these tasks. To conduct this investigation, we utilize LLMs, specifically ChatGPT and PaLM2, as paraphrasers, alongside other paraphrasing models (PMs) such as Pegasus \citep{zhang2020pegasus}, operating at the sentence level, and Dipper \citep{krishna2024paraphrasing}, capable of whole-text paraphrasing while preserving contextual coherence and providing control over lexical diversity. Our comprehensive analysis encompasses diverse text sources, including human-authored content and texts generated by six LLMs, paraphrased by these four paraphrasers across seven distinct datasets.
}


Our study stands apart from other research in authorship analysis, paraphrasing detection, AI-text detection, or style analysis. The major contribution of our paper is as follows:
\begin{itemize}
\item \textcolor{black}{ We explore the counter-intuitive assumptions associated with paraphrasing and authorship by adopting a comprehensive computational perspective, drawing inspiration from the Ship of Theseus scenario.}
    \item We identify the difference among paraphrasers regarding their effect on authorship.
    \item We create a paraphrased corpus\footnote{Available at \url{https://github.com/tripto03/Ship_of_theseus_paraphrased_copus}} consisting of seven sources (six LLMs+human-authored), seven datasets, and four paraphrasers. 
\end{itemize}

\textcolor{black}{
In the traditional scenario, the classification models exhibit a substantial drop in performance after the initial paraphrase compared to the original texts, with diminishing declines for subsequent paraphrases. Conversely, in the alternative scenario, the classifier's performance experiences a milder decrease. This phenomenon indicates that LLMs imprint their distinctive "style" onto the paraphrased texts. However, conceptualizing the ground truth of authorship should be contingent upon the specific task, a point we substantiate through the lens of relevant philosophical and writing theories.
}

 
\section{Related Work}
\label{sec-related_works}

\begin{table*}[!htb]
\renewcommand{\tabcolsep}{2pt}
\centering
\footnotesize
\resizebox{\textwidth}{!}{
\begin{tabular}{@{}|l|l|l|@{}}
\toprule
\multicolumn{1}{|c|}{\textbf{Dataset with Sample Size}}                                                                                                & \multicolumn{1}{c|}{\textbf{Authors: Organization}}                                                    & \multicolumn{1}{c|}{\textbf{Paraphraser}}                                                                                                                                                                                                                                                                                                                                                                                 \\ \midrule
\begin{tabular}[c]{@{}l@{}}\textbf{Xsum} \citep{narayan2018xsum}: 956 \\ news articles in various topics\end{tabular}                      & \begin{tabular}[c]{@{}l@{}}\textbf{Human}: original source of\\  writings\end{tabular}        & \multirow{2}{*}{\begin{tabular}[c]{@{}l@{}}\textbf{ChatGPT}: we utilize the prompt "\textit{paraphrase the} \\ \textit{following text. keep similar length}" to paraphrase any \\ given text. We set the max length as the allowed max\\ length and keep the other parameters as default.\end{tabular}}                                                                                                                         \\ \cmidrule(r){1-2}
\begin{tabular}[c]{@{}l@{}}\textbf{TLDR}: 766 articles collected from\\ daily tech newsletter \tablefootnote{\url{https://huggingface.co/datasets/JulesBelveze/tldr_news}}\end{tabular}                                & \begin{tabular}[c]{@{}l@{}}\textbf{ChatGPT} (\textit{gpt 3.5 turbo}):\\ OpenAI\end{tabular}          &                                                                                                                                                                                                                                                                                                                                                                                                      \\ \midrule
\begin{tabular}[c]{@{}l@{}}\textbf{SCI}\_\textbf{GEN} \citep{moosavi2021sci_gen}\\ 944 abstracts of scientific articles \end{tabular}                                     & \begin{tabular}[c]{@{}l@{}}\textbf{PaLM2} (\textit{text-bison@001})\\ \citep{anil2023palm2}: Google \tablefootnote{Original FLAN-T5 samples in \citep{li2023deepfake} were substantially smaller compared to other LLMs, so we replace them by generating text from PaLM2 with same prompt}\end{tabular}          & \begin{tabular}[c]{@{}l@{}}\textbf{PaLM2}: similar technique. PaLM2 often generates text \\ with formatting that we remove to keep the plain text only\end{tabular}                                                                                                                                                                                                                                                \\ \midrule
\begin{tabular}[c]{@{}l@{}}\textbf{CMV} \citep{tan2016CMV}: total 514 \\ statements from r/ChangeMyView \\ SubReddit\end{tabular}         & \begin{tabular}[c]{@{}l@{}}\textbf{LLaMA}-65B \cite{touvron2023llama}:\\ 
Meta\end{tabular}     & \multirow{2}{*}{\begin{tabular}[c]{@{}l@{}}\textbf{Dipper} \citep{krishna2024paraphrasing}: can paraphrase the whole\\ text by controlling output diversity. We consider lexical\_\\ diversity(\textit{lex}) = 60, order\_diversity(\textit{order})=60 as the \\ default \textbf{dipper(moderate)} setting. We perform ablation \\ with \textbf{dipper(high)} settings as \textit{lex=100,order=100} and \\ \textbf{dipper(low)} setting as \textit{lex=20,order=20}.\end{tabular}} \\ \\ \cmidrule(r){1-2}
\begin{tabular}[c]{@{}l@{}}\textbf{WP} \citep{fan2018writing_prompt}: 942 stories\\ based on prompts from \\ r/WritingPrompts SubReddit\end{tabular} & \begin{tabular}[c]{@{}l@{}}GLM-130B \cite{zeng2022glm130B}:
\\ \textbf{Tshinghua}\end{tabular}  &                                                                                                                                                                                                                                                                                                                                                                                                      \\ \midrule
\begin{tabular}[c]{@{}l@{}}\textbf{ELI5} \citep{fan2018writing_prompt}: 954 answers\\ from r/ExplainLikeIam5 SubReddit\end{tabular}                  & \begin{tabular}[c]{@{}l@{}}\textbf{BLOOM}-7B1 \cite{scao2022bloom}:\\
BigScience\end{tabular} & \multirow{2}{*}{\begin{tabular}[c]{@{}l@{}}\textbf{Pegasus} \citep{zhang2020pegasus}: a sentence-wise paraphraser.\\  We paraphrase all sentences in a text as default setting.\\ We perform ablation study with \textbf{pegasus(slight)} variation\\ that paraphrases random 25\% sentences in a text.\end{tabular}}                                                                                                      \\ \cmidrule(r){1-2}
\begin{tabular}[c]{@{}l@{}}\textbf{YELP} \citep{zhang2015yelp}: 986 \\ reviews from yelp dataset\end{tabular}                               & \begin{tabular}[c]{@{}l@{}}GPT-NeoX-20B \cite{black2022gpt_neox}: \\ 
\textbf{EleutherAI}\end{tabular}                &                                                                                                                                                                                                                                                                                                                                                                                                      \\ \bottomrule
\end{tabular}
}
\caption{Summary of datasets, authors, and paraphrasers. ChatGPT and PaLM2 serve as both candidate authors for text generation and paraphrasers as well. Sample size indicates the original human writings that were considered for each dataset. For instance, \textbf{xsum} dataset will contain approximately 956$\times$50\%(test split)$\times$7(authors)$\times$4(paraphrasers)$\times$3(times paraphrasing) $\approx$  40K samples. }
\label{tab-dataset_table}
\end{table*}

\textcolor{black}{
Our study extends prior research on authorship attribution tasks from various perspectives, including style and content. Notably, \citet{sari2018topic_or_style} found that content-based features are more effective for datasets with high topical variance, while datasets with lower variance benefit more from style-based features for these tasks.
}
Several assessments and benchmarks on stylistic analysis have aimed to identify and infer style across different domains. The XSLUE benchmark \citep{kang2021xslue} comprehensively evaluates sentence-level cross-style language understanding in 15 different styles. Additionally, the STEL framework \citep{wegmann-nguyen-2021-stel, wegmann2022same} introduces four specific assessments measuring the stylistic content of authorship representations: formality, simplicity, contraction usage, and number substitution preference.
Recent research has also explored the learning of authorship representations \citep{boenninghoff2019explainable,hay2020representation} in diverse cross-domain settings. For instance, \citet{rivera2021learning} introduced the concept of universal authorship representations with a recent extension \citep{wang2023can_universal_authorship} to validate their capacity to capture stylistic features. However, it is essential to note that these studies primarily focus on performing classification tasks related to authorship in various setups. Our task distinguishes itself by delving into the established concept of ground truth concerning authorship in paraphrasing in the era of LLMs.

Another body of related research revolves around paraphrasing detection and plagiarism. These studies aim to determine whether a pair of texts constitutes a paraphrased version of one another \citep{becker2023paraphrase}. Paraphrasing detection stands as a critical challenge within the domain of plagiarism identification \citep{chowdhury2018plagiarism}. It is also a subject of inquiry in evaluating a proposed model's capacity for addressing natural language understanding tasks \citep{wang2018glue}. Previous research encompasses both human-generated \citep{seraj2015improving, dong2021parasci} and machine-generated \citep{foltynek2020detecting, wahle2022identifying} paraphrased versions. A recent investigation by \citet{wahle2022large} suggests that machine-generated paraphrases bear greater similarity to the original source text than human-generated paraphrases. This phenomenon resurges the discussion: if an LLM, such as ChatGPT, is employed to paraphrase a text, should ChatGPT be regarded as the author? Therefore, our study seeks to explore the connection between authorship and paraphrasing by bridging the gap among these distinct lines of research.
\begin{figure*}[h]
    \centering
    \includegraphics[width=0.8\linewidth]{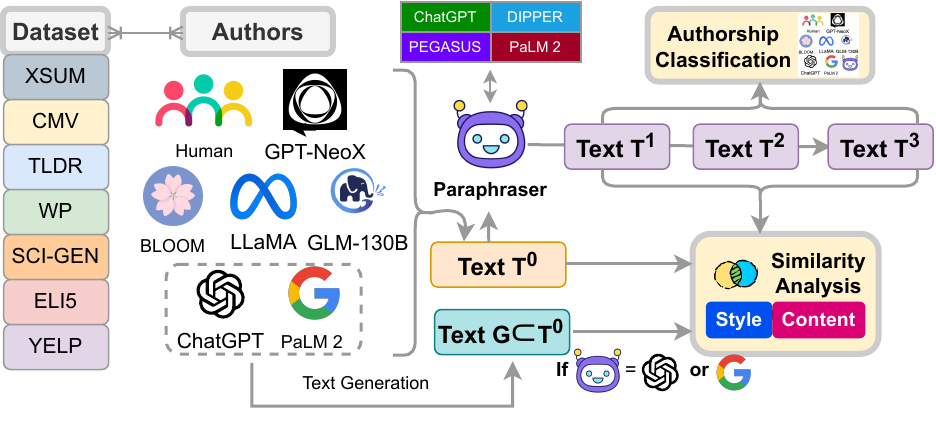}
    \caption{The original benchmark from \citet{li2023deepfake} has several datasets with samples from different sources (human and LLMs) in each dataset. Original texts ($T^0$) are paraphrased sequentially three times, utilizing diverse paraphrasers (LLMs or PMs). We assess classifier performance in each iteration and measure their resemblance to $T^0$. For LLM paraphrasers, we additionally evaluate their similarity with text $G$, generated by the respective LLM.}
    \label{fig_methods}
\end{figure*}

\section{Methodology}
\label{sec_method}
The classical authorship attribution problem aims to determine the author ($A$) of a given text $T^0$ from a set of candidate authors, typically treated as a multi-class classification task. However, when text $T^0$ is paraphrased into $T^1$ by an LLM ($B$), who is also a potential author in the candidate set, it raises a question of what should be considered the ground truth. The \textbf{traditional} perspective designates the original author $A$ as the author of $T^1$, while an \textbf{alternative} perspective assigns LLM $B$ as the author. Text $T^1$ may substantially diverge from $T^0$ in style and content, potentially more similar to text $G$, independently generated by LLM $B$ on the same subject. A similar scenario is also applicable to human vs. AI text detection problems. Therefore, our methodology (Figure \ref{fig_methods}) focuses on assessing the classifier's performance, considering both perspectives and exploring how variations in style and content account for the observed differences.

\paragraph{Dataset development:}

We built our dataset from the benchmark by \citet{li2023deepfake}, which features text generated by various LLMs using the same prompt, specifically instructing LLMs to continue generating text based on the first 30 words of the original human-written text. This choice enables us to explore a realistic scenario where multiple authors have written on the same subject. Given the remarkable similarity between recent LLM-generated (AI) text and human text, it is not meaningful to classify authors at the single-sentence level \citep{yang2023survey}. Therefore, we selected seven datasets with paragraph-level texts from \citet{li2023deepfake} and included one LLM from each language model family. Table \ref{tab-dataset_table} provides a concise overview of these datasets, the selected authors, and the paraphrasers (LLM or PM).

Each dataset is divided into a 50:50 split, allocating half for training classifiers and constructing style models and the other half for paraphrasing evaluations. {\ul To prevent the classifiers from being exclusively trained on the text's topic or content, we ensured that the train and test portions contain identical split (based on originating source) from all authors}. We paraphrased each text in test portion three times, sequentially, i.e. the original text $T$ is paraphrased once to obtain text $T^1$, which is then paraphrased again to get $T^2$ and then once again to generate $T^3$.



\paragraph{Author style model:}
While specific contrastive learning-based techniques \citep{wegmann-nguyen-2021-stel,wang2023can_universal_authorship} aim to discern context-independent style embeddings, we opted not to employ them as our style model. Firstly, these techniques operate as black-box embeddings, making it challenging to comprehend their inner workings and ensure explainability \citep{angelov2021explainable}. Additionally, our aim to validate whether the drop in classification performance can be attributed to changes in style, employing a more accessible perspective. Hence, we utilized a feature-based approach, incorporating features from LIWC \citep{pennebaker2001LIWC}, and WritePrints \citep{abbasi2008writeprints_DT}, 
to construct our style model for each author in individual datasets.

\textcolor{black}{
We define the style model derived from each author's original test samples ($T^0$) as its baseline and assess its deviation from the styles in $T^1$, $T^2$, and $T^3$, respectively. We employ the remaining training portion for each author to validate its applicability. A robust style model should yield similar results between the style models from the training and original test samples ($T^0$). Mathematically, for authors A and B with respective training and test style models, their distances should adhere to the following properties:}
\[
\begin{aligned}
\lvert \text{train}(A) - \text{test}(A) \rvert &< \lvert \text{train}(B) - \text{test}(A) \rvert \\
\lvert \text{train}(B) - \text{test}(B) \rvert &< \lvert \text{train}(A) - \text{test}(B) \rvert
\end{aligned}
\]
\textcolor{black}{
We validate this behavior using Mahalanobis distance (MD) \citep{mclachlan1999mahalanobis}, measuring the distance between a point and a distribution. For instance, to validate $\lvert \text{train}(A) - \text{test}(A) \rvert < \lvert \text{train}(B) - \text{test}(A) \rvert$, we calculate the Mahalanobis distance $M$$D(x, \text{train}(A))$ and $M$$D(x, \text{train}(B))$ for each point $x \in test(A)$. We utilize a one-sample Wilcoxon signed-rank test \citep{conover1999wilscon_ttest} to demonstrate that $M$$D(x, \text{train}(A)) < M$$D(x, \text{train}(B))$ is statistically significant for any given author A and B. The results consistently yield a p-value $<0.001$ across all datasets and authors. The performance of the style model as a classifier also validates its effectiveness in classifying text in its original state (details in Appendix \ref{subsec_style_model_validity}).
}

\paragraph{Content similarity:}
We employ \textit{text-embedding-ada-002} from OpenAI to assess the deviation of paraphrased text from the original text's content. It is known for its high performance in the Massive Text Embedding Benchmark (MTEB) leaderboard \citep{muennighoff2023mteb} and can take lengthy texts as input (up to 8191 tokens) compared to others. Our analysis also establishes its correlation with pairwise BERT \citep{zhang2019bertscore} and BLEU \citep{papineni2002bleu} scores between original and paraphrased texts.






\section{Experimental Results}

We primarily focus on evaluating the impact of paraphrasing in the context of authorship attribution, which translates into a seven-class classification problem. 
Our objective is not to devise new text classification methods but to investigate how ground truth influences their performance and its correlation with changes in style and content. We employ established text classification methods, including \textbf{Finetuned BERT} \textit{(bert-base-cased)} as a representative of the finetuned language model (LM), our style model with XGBoost \citep{chen2016xgboost} classifier as a representation of \textbf{stylometry}, \textbf{GPT-who} \citep{venkatraman2023gptwho} for information density-based multi-class classification, and \textbf{TF-IDF} with logistic regression for classic n-gram-based analysis. We also explore the human vs. AI text detection scenario, a binary classification problem, using different finetuned and zero shot approaches. 


\begin{table}[h]
\centering
\footnotesize
\resizebox{0.48\textwidth}{!}{
\begin{tabular}{@{}p{1.5cm}@{}lllll@{}}
\toprule
\multicolumn{2}{c}{\textbf{Text $T^n$}} & \textbf{BERT} & \textbf{Stylometry} & \textbf{GPT-who} & \textbf{TF-IDF} \\
\midrule
\textbf{Original} & \textbf{0} & \textbf{0.77} & 0.71 & 0.62 & 0.61 \\
\midrule
\multirow{3}{*}{\textbf{ChatGPT}} & \textbf{1} &  0.33 \dec{57.1} & 0.32 \dec{54.9} & 0.28 \dec{54.8} & \textbf{0.35} \dec{42.6}  \\
& \textbf{2} &  0.31 \dec{6.1} & 0.29 \dec{9.4} & 0.27 \dec{3.6} & \textbf{0.33} \dec{5.7}  \\
& \textbf{3} &  0.29 \dec{6.5} & 0.27 \dec{6.9} & 0.25 \dec{7.4} & \textbf{0.32} \dec{3.0} \\
\midrule
\multirow{3}{*}{\textbf{PaLM2}} & \textbf{1} & \textbf{0.44} \dec{42.9} & 0.43 \dec{39.4} & 0.35 \dec{43.5} & 0.43 \dec{29.5} \\
& \textbf{2} & 0.39 \dec{11.4} & 0.38 \dec{11.6} & 0.32 \dec{8.6} & \textbf{0.4} \dec{7.0}  \\
& \textbf{3} &  0.37 \dec{5.1} & 0.37 \dec{2.6} & 0.3 \dec{6.3} & \textbf{0.39} \dec{2.5} \\
\midrule
\multirow{3}{*}{\textbf{Dipper}} & \textbf{1} &  0.38 \dec{50.6} & 0.35 \dec{50.7} & 0.33 \dec{46.8} & \textbf{0.44} \dec{27.9}  \\
& \textbf{2} &  0.33 \dec{13.2} & 0.31 \dec{11.4} & 0.28 \dec{15.2} & \textbf{0.38} \dec{13.6}  \\
& \textbf{3} &  0.29 \dec{12.1} & 0.29 \dec{6.5} & 0.25 \dec{10.7} & \textbf{0.35} \dec{7.9} \\
\midrule
\multirow{3}{*}{\textbf{Pegasus}} & \textbf{1} &  \textbf{0.55} \dec{28.6} & 0.49 \dec{31.0} & 0.44 \dec{29.0} & 0.49 \dec{19.7}  \\
& \textbf{2} &  \textbf{0.49} \dec{10.9} & 0.46 \dec{6.1} & 0.4 \dec{9.1} & 0.45 \dec{8.2} \\
& \textbf{3} &  \textbf{0.47} \dec{4.1} & 0.42 \dec{8.7} & 0.38 \dec{5.0} & 0.42 \dec{6.7}  \\
\bottomrule
\end{tabular}
}
\caption{Performance (avg. of macro f1 score across datasets) of classifiers for various paraphrasers across different versions (\textbf{traditional} perspective). \textcolor{red}{$\downarrow$} denotes performance drop from the previous version $T^{n-1}$.}
\label{tab:AA_results}
\end{table}

\begin{figure*}[h]
  \begin{subfigure}{0.32\textwidth}
    \centering
    \includegraphics[width=\linewidth]{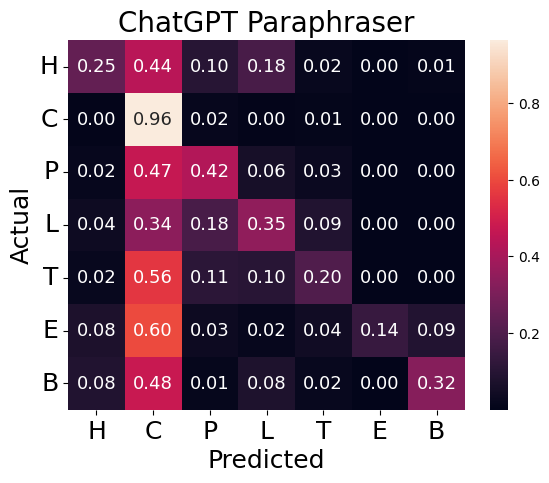}
  \end{subfigure}
    \hfill
  \begin{subfigure}{0.32\textwidth}
    \centering
    \includegraphics[width=\linewidth]{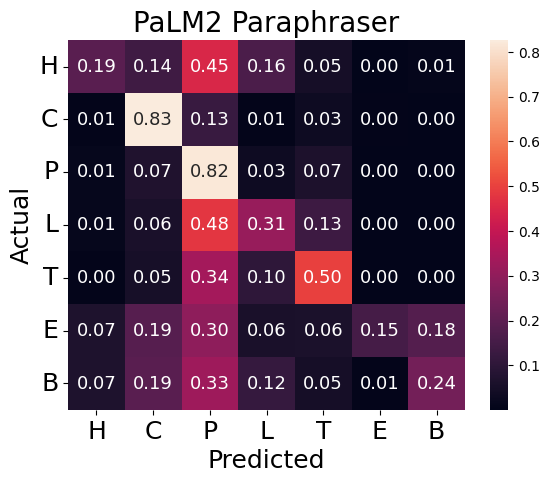}
  \end{subfigure}
    \hfill
  \begin{subfigure}{0.32\textwidth}
    \centering
    \includegraphics[width=\linewidth]{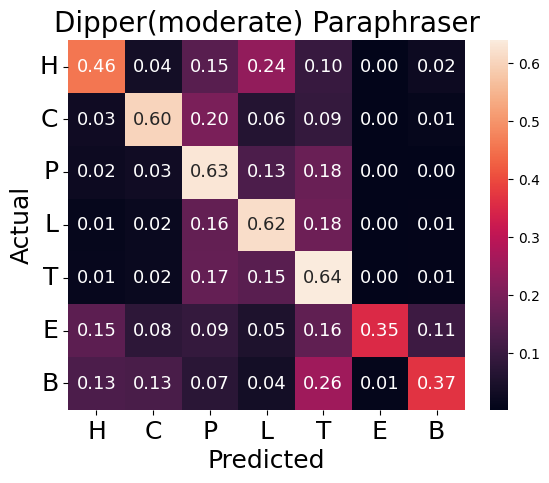}
  \end{subfigure}
  \caption{Confusion matrix for the Fine-tuned BERT classifier after the first version of paraphrasing (H: Human, C: ChatGPT, P: PaLM2, L: LLAMA, T: Tsinghua, E: Eleuther-AI, B: Bloom). In the case of LLM paraphrasers (ChatGPT and PaLM2), paraphrased samples are predominantly misclassified as corresponding LLMs, whereas for other PMs, such as Dipper (and Pegasus also), misclassifications are distributed more uniformly.}
  \label{fig-confusion-matrix}
\end{figure*}

\begin{table*}[!htb]
\renewcommand{\tabcolsep}{2pt}
\centering
\footnotesize
\resizebox{\textwidth}{!}{
\begin{tabular}{@{}l|cc|cc@{}}
\toprule
\textbf{Paraphraser}  & \multicolumn{2}{c|}{\textbf{ChatGPT}}                            & \multicolumn{2}{c}{\textbf{PaLM2}}                              \\ \midrule
\textbf{Ground truth} & \multicolumn{1}{c|}{\textbf{traditional}} & \textbf{alternative} & \multicolumn{1}{c|}{\textbf{traditional}} & \textbf{alternative} \\ \midrule
\textbf{xsum}         & \multicolumn{1}{l|}{0.72→0.26→0.24→0.22}  & 0.72→0.65→0.65→0.66  & \multicolumn{1}{l|}{0.72→0.45→0.4→0.38}   & 0.72→0.62→0.63→0.66  \\ \midrule
\textbf{cmv}          & \multicolumn{1}{l|}{0.82→0.32→0.28→0.26}  & 0.82→0.75→0.76→0.71  & \multicolumn{1}{l|}{0.82→0.38→0.35→0.33}  & 0.82→0.71→0.7→0.71   \\ \midrule
\textbf{sci\_gen}     & \multicolumn{1}{l|}{0.77→0.43→0.39→0.35}  & 0.71→0.69→0.71→0.71  & \multicolumn{1}{l|}{0.77→0.55→0.51→0.5}   & 0.77→0.68→0.69→0.69  \\ \bottomrule
\end{tabular}
}
\caption{Performance comparison between \textbf{traditional} and \textbf{alternative} perspectives of ground truth in subsequent LLM paraphrasing ($T^0$ → ... → $T^3$). The values indicate the F1 score for authorship attribution for the finetuned BERT classifier (best-performing method).}
\label{tab_comparision_table}
\end{table*}

\paragraph{Authorship attribution results:}
Table \ref{tab:AA_results} presents the impact of different paraphrasing iterations on classifier performance in the \textbf{traditional perspective} of ground truth. A notable performance drop is observed after the first paraphrased version (from original $T^0$ to $T^1$), with subsequent iterations causing marginal decreases in all cases. ChatGPT paraphrasers exhibit the most substantial performance drop, while Pegasus has the slightest effect. Additionally, it is interesting to note the performance variation among classifiers. BERT, which achieves the highest performance in original datasets ($T^0$), is the most affected by paraphrasing, followed by stylometry. In contrast, TF-IDF, initially the lowest performer in the original dataset, exhibits the highest F1 score when dealing with paraphrased text, except for Pegasus. This suggests that paraphrasing primarily impacts style while retaining similar content, making it challenging for classifiers to attribute samples accurately. Figure \ref{fig-confusion-matrix} delves deeper into the causes of performance drops and misclassifications of authors after paraphrasing.

\textcolor{black}{
Table \ref{tab_comparision_table} presents the results of authorship attribution under the \textbf{alternative perspective}, wherein authorship is considered to change after paraphrasing, in contrast to the traditional perspective across various datasets. Notably, the classification model exhibits significantly higher performance under this alternative perspective compared to the traditional perspective after the first iteration. For instance, there is an average performance drop of 57.1\% in the traditional perspective versus only a 6.9\% drop in the alternative perspective for ChatGPT paraphrase. Moreover, for subsequent paraphrasing iterations, while the performance drop is marginal in both perspectives, it remains higher in the traditional scenario compared to the alternative perspective. This difference is also statistically significant, as indicated by the Wilcoxon signed-rank test (p < 0.05).
Although adopting the alternative perspective as the ground truth in all scenarios may appear advantageous, it is not universally applicable, as we discuss in the following section.
}

\begin{figure*}
  \begin{subfigure}{0.32\textwidth}
    \centering
    \includegraphics[width=\linewidth]{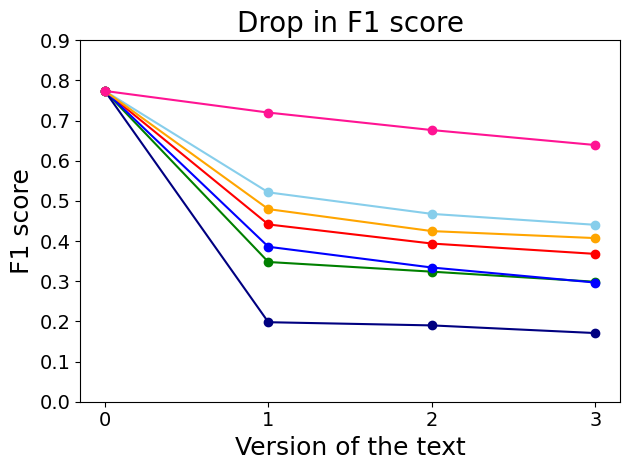}
  \end{subfigure}
    \hfill
  \begin{subfigure}{0.32\textwidth}
    \centering
    \includegraphics[width=\linewidth]{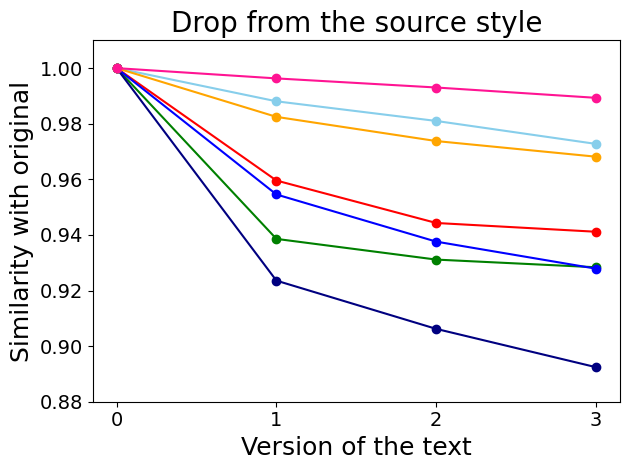}
  \end{subfigure}
    \hfill
  \begin{subfigure}{0.32\textwidth}
    \centering
    \includegraphics[width=\linewidth]{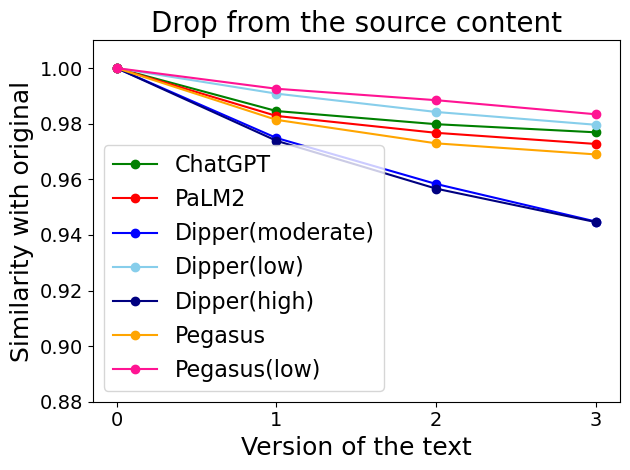}
  \end{subfigure}
  \caption{Comparison of classification performance (avg. macro F1 score for Fine-tuned BERT), style deviation from the original, and content shift in successive paraphrased versions across various paraphrasing methods (averaged across all datasets and sources). Style and content deviations are calculated as the average of the cosine distance between the corresponding feature vector (author style model and \textit{text-embedding-ada-002} respectively) between paraphrased text $T^n$ and the original text $T^0$.}
  \label{fig-drop-from-source}
\end{figure*}
\begin{figure}[h]
  \begin{subfigure}[b]{0.48\columnwidth}
    \includegraphics[width=\linewidth]{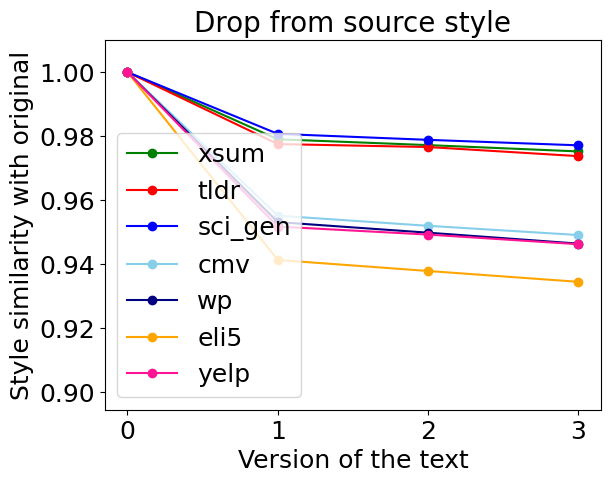}
  \end{subfigure}
  \hfill 
  \begin{subfigure}[b]{0.48\columnwidth}
    \includegraphics[width=\linewidth]{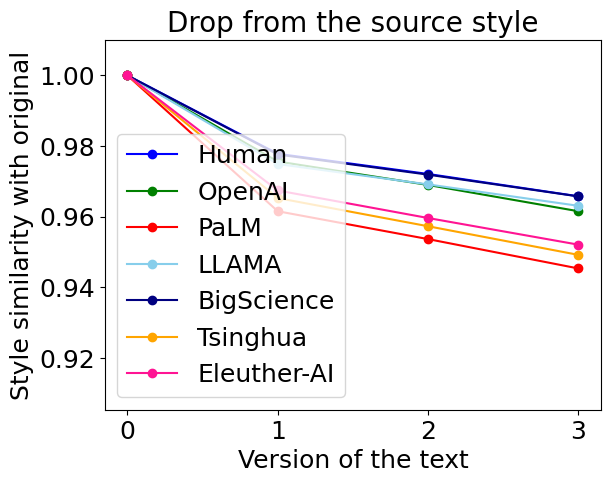}
  \end{subfigure}
  \caption{How style similarity drops from original text for different authors and different datasets.}
  \label{fig_drop_in_dataset_author}
\end{figure}

\paragraph{Style and content similarity drop:}
For each paraphrase, we examine how the paraphrased text deviates from the original text across two crucial dimensions: \textbf{content} and \textbf{style}. 
Figure \ref{fig-drop-from-source} illustrates the reduction in style and content similarity between the original text and its paraphrased versions, with the performance drop. We note that style deviates more substantially than content, with a substantial drop after the first paraphrasing and marginal changes in subsequent iterations, resembling the F1 score trend for all paraphrasers. A Pearson correlation test confirmed the statistically significant relation (p < 0.05) between the decline in F1 scores and a drop in style similarity. Figure \ref{fig_drop_in_dataset_author} shows the individual breakdown of style drop for different authors and datasets.

\begin{figure}[h]
  \begin{subfigure}[b]{0.48\columnwidth}
    \includegraphics[width=\linewidth]{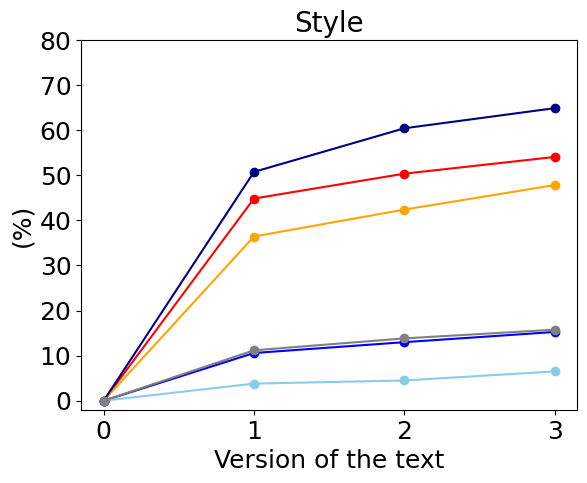}
  \end{subfigure}
  \hfill 
  \begin{subfigure}[b]{0.48\columnwidth}
    \includegraphics[width=\linewidth]{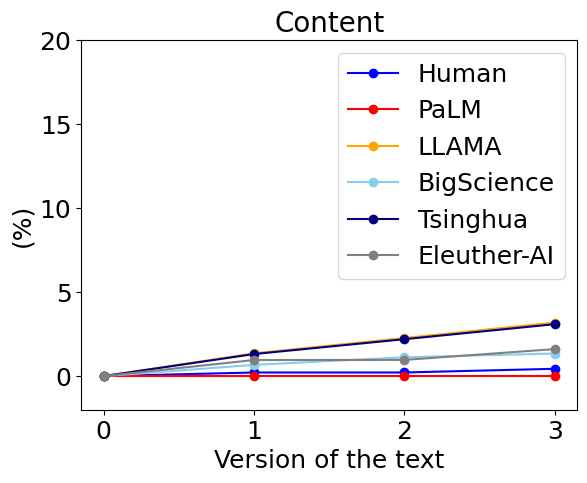}
  \end{subfigure}
  \caption{Percentages of ChatGPT paraphrased text ($T^n$) more similar to ChatGPT-generated original text ($G$) than the original authors' text ($T^0$) in style/content.}
  \label{fig_llm_similarity}
\end{figure}

Figure \ref{fig_llm_similarity} illustrates that for specific authors, following successive paraphrasing, a substantial portion of LLM paraphrased text can become more similar to the style of the LLM.



\paragraph{AI text detection results:} 
Using four zero-shot AI text detectors (DetectGPT \citealp{mitchell2023detectgpt}, GPTZero \citealp{GPTzero}, RoBERTa, and LongFormer \citealp{li2023deepfake}) and fine-tuned BERT, we assess current AI text detectors in three scenarios. As expected, fine-tuned BERT surpasses all zero-shot approaches. LongFormer exhibits the best performance among the zero-shot methods, likely due to their fine-tuning dataset's overlap with ours. 

The detection of AI-generated text is easiest in the original setting as compared to both scenarios involving paraphrased texts as can be seen (Table \ref{tab:AI_detection_results}) from the higher performance in the \textbf{original} setting across both datasets. For scenarios with paraphrased text, AI text detectors perform better under the \textbf{traditional} scenario i.e. when paraphrasing does not change authorship than in the \textbf{alternative} scenario i.e. when paraphrased text is assumed to be authored by the paraphraser for formal writing styles (\textbf{xsum)}, while the opposite is true for most detectors in the informal writing task (\textbf{eli5}). 

\begin{table}[h]
  \centering
  \resizebox{\columnwidth}{!}{
    \begin{tabular}{l|l|lllll}
      \toprule
      \textbf{Dataset} & \textbf{Scenario} & \textbf{BERT } & \textbf{Detect-} & \textbf{GPT-} & \textbf{RoBERTa} & \textbf{Long- } \\
      & & (finetuned) &\textbf{GPT} &\textbf{Zero} &  (zero-shot)& \textbf{Former} \\
      \midrule
      \multirow{3}{*}{\textbf{xsum}} & original & \textbf{0.97} & 0.66 & 0.67 & 0.3 & 0.9 \\
                                     & traditional & \textbf{0.83} & 0.66 & 0.66 & 0.36 & 0.68 \\
                                     & alternative & 0.6 & \textbf{0.67} & \textbf{0.67} & 0.33 & 0.36 \\
      \midrule
      \multirow{3}{*}{\textbf{eli5}} & original & \textbf{0.96} & 0.67 & 0.71 & 0.34 & 0.86 \\
                                     & traditional & 0.51 & 0.66 & 0.66 & 0.39 & \textbf{0.69} \\
                                     & alternative & \textbf{0.88} & 0.67 & 0.75 & 0.33 & 0.4 \\

      \bottomrule
    \end{tabular}
  }
    \caption{Performance of AI text detectors along different authorship perspectives after ChatGPT paraphrasing.}
  \label{tab:AI_detection_results}
\end{table}

\begin{figure*}[h]
  \begin{subfigure}{0.49\textwidth}
    \centering
    \includegraphics[height=3.5cm,width=\linewidth]{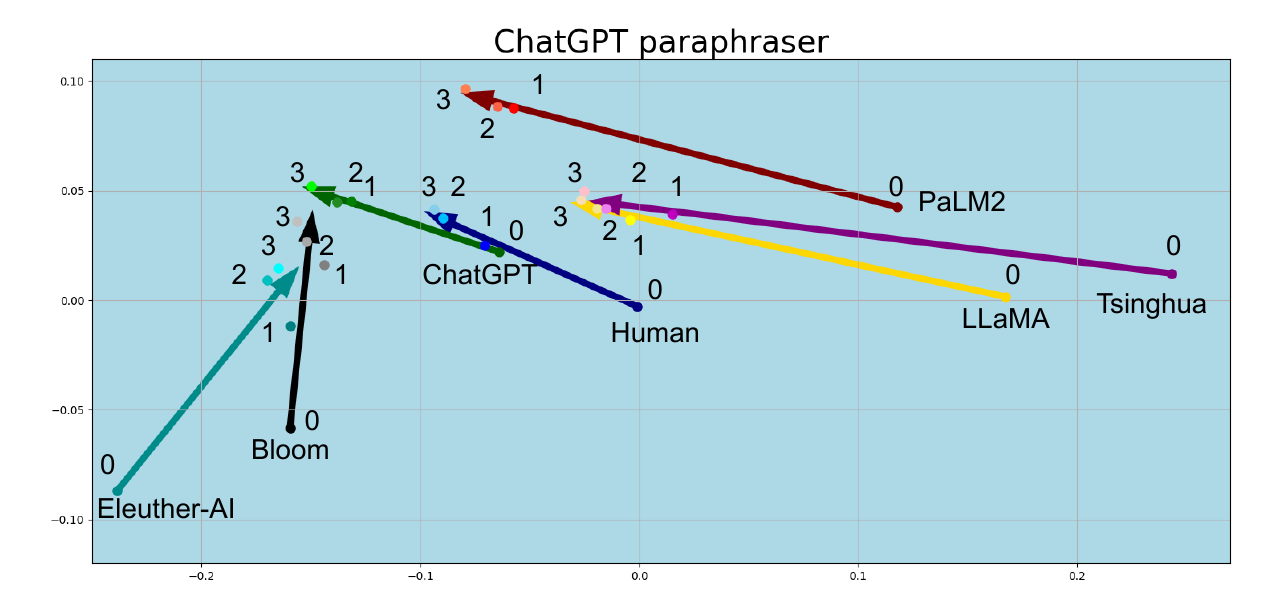}
  \end{subfigure}
    \hfill
  \begin{subfigure}{0.49\textwidth}
    \centering
    \includegraphics[height=3.5cm,width=\linewidth]{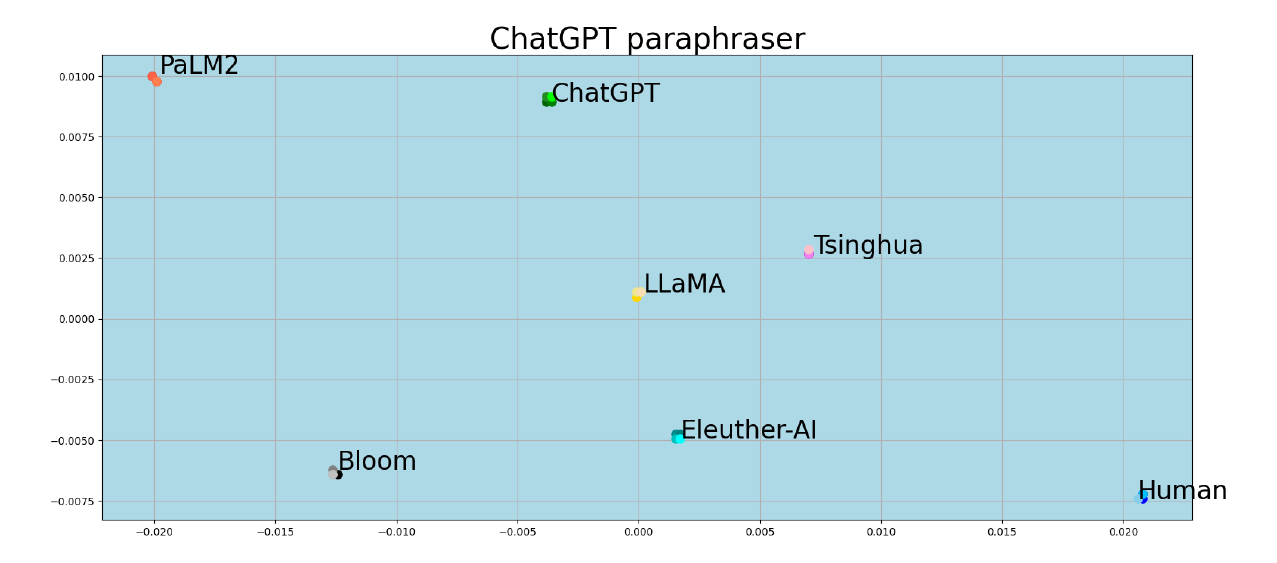}
  \end{subfigure}
  \caption{PCA visualization of author style (left) and content (right) shifts in \textbf{cmv} dataset for ChatGPT paraphraser. While style shifts substantially, the content of paraphrased versions remains same (details in Figure \ref{fig_pca_visualization}, Appendix).}
  \label{fig_content_style_shift}
\end{figure*}

\section{Discussion}
\subsection{Major Findings}
We discuss the significant findings from results that contribute essential insights to our subsequent discussion regarding paraphrasing and authorship.
\paragraph{Style deviates a lot more than content after paraphrasing:}
Our  observation highlights a substantial divergence in style compared to content when paraphrasing is applied.
PMs and LLMs seek to preserve the original semantics of the text while paraphrasing, and {\ul this deviation from style is why classifiers fail to correctly attribute the paraphrased versions}.

\paragraph{LLM paraphrasers deviate the style to the LLM style model:}
LLM paraphrasers tend to align the paraphrased text's style with the LLM's style model (Figure \ref{fig_pca_visualization}). The paraphrased text also often exhibits greater stylistic similarity to the LLM's original text than the actual author (Figure \ref{fig_llm_similarity}). This explains the misclassification of paraphrased texts as the corresponding LLM label (Figure \ref{fig-confusion-matrix}).


\paragraph{Subsequent paraphrasing differs for LLM and PM paraphrasers:}
Our findings reveal a notable performance drop and style deviation following the first paraphrasing. Though, subsequently we note a mere 3\%-4\% average performance decrease from the second paraphrase onward, signifying less impact. In contrast, {\ul  PM paraphrasers display a consistent decline in style compared to LLM paraphrasers in subsequent versions} ($T^1$ to $T^2$ and $T^2$ to $T^3$). 

\paragraph{The choice of paraphraser impacts performance and style deviation:}
{\ul ChatGPT is a stronger paraphraser than PaLM2} as per both performance drop and style deviation. Additionally, our observations highlight the significance of lexical diversity, as evidenced by the variations in Dipper's low, moderate, and high versions.

\paragraph{Performance varies across datasets and sources:}
{\ul The impact of paraphrasing is milder for formal writing datasets} like \textbf{xsum}, \textbf{tldr}, and \textbf{scigen} versus informal ones, such as \textbf{eli5}, \textbf{cmv}, and \textbf{yelp}. Formal writings, such as scientific abstracts and news articles, often exhibit a consistent style across sources. In contrast, informal writings, such as Reddit comments or reviews, have greater style diversity and variance. Thus, paraphrasing substantially alters the style of informal text.
Also, {\ul Human and ChatGPT-generated texts maintain their original style through paraphrasing iterations}, while PaLM2 and Tsinghua texts undergo substantial stylistic changes after paraphrasing.

\subsection{Philosophical Perspective of Authorship}
The original Ship of Theseus paradox that motivates our research remains a topic of philosophical debate without universal consensus \citep{pickup2016situationalist_solution}. 
\textcolor{black}{
For example, Identity through Continuity \citep{wiggins1967identity_through_continuity,cohen2004identity_ship} and Bundle Theory suggest \citep{pike1967hume_bundle_theory} that the identity of an object is tied to its continuity or the persistence of specific essential characteristics that it possesses. Applied to the Ship of Theseus paradox, this means that as long as the ship's fundamental attributes like design, purpose, and function remain unchanged despite part replacements, it can still be considered the same. Drawing a parallel connection in the context of authorship, the Originality of Expression theory \citep{samuels1988idea,foucault2003author}  argues that authorship is based on the author's unique ideas, concepts, or thoughts. If a paraphraser modifies the expression while preserving the core ideas, the original authorship may still be attributed to the original author. Our investigation, however, shows that substantial stylistic variation from the source due to paraphrasing can reduce classification performance. While some argue for Transformational Authorship \citep{sarja2023transformative}, in which authorship can be attributed to those who transform a text, even through paraphrasing, it is essential to note that the paraphraser's style is ultimately rooted in samples generated by humans or other entities, and this style is not fixed but can be controlled through various parameters and prompting. 
}

Table \ref{tab-ship_solution} summarizes these potential solution scenarios, drawing parallels to authorship scenarios and supporting theories. The notion of authorship is multifaceted and context-sensitive. Whether LLM paraphrasing should alter authorship depends on the use cases discussed below.

\renewcommand{\tabcolsep}{2pt}
\begin{table*}[h]
\centering
\footnotesize
\resizebox{\textwidth}{!}{
\begin{tabular}{@{}|l|l|l|l|l|@{}}
\toprule
\multicolumn{1}{|c|}{\textbf{Ship scenario}}                                                                                                & \multicolumn{1}{|c|}{\textbf{Supporting Philosophical Theory}}                                                                                                                                                                                                                                                                                                                                              & \multicolumn{1}{|c|}{\textbf{Authorship Scenario}}                                                                                     & \multicolumn{1}{|c|}{\textbf{Supporting Writing Theory}}                                                                                                                                                                                                                                                                                                                                         & \multicolumn{1}{|c|}{\textbf{Applicability}}                                                                                     \\ \midrule
\begin{tabular}[c]{@{}l@{}}The original ship with \\ \textit{planks replaced} should \\ be considered original\end{tabular}     & \begin{tabular}[c]{@{}l@{}}\textbf{Bundle theory} \citep{pike1967hume_bundle_theory}: Identity\\ of an object is tied to the persistent \\ of the specific characteristics that it\\ present. As long as the ship's bundle \\ of properties/characteristics remain\\ unchanged, it is the same ship.\end{tabular}                                                                                                    & \begin{tabular}[c]{@{}l@{}}Paraphrasing \textbf{should}\\ \textbf{not change} authorship\end{tabular}                              & \begin{tabular}[c]{@{}l@{}}\textbf{The idea-expression dichotomy} \\ \citep{samuels1988idea}: Authorship is based \\ on the author's unique expression of \\ ideas, concepts, or thoughts. If a \\ paraphraser modifies the expression \\ while preserving the core ideas, the \\ original authorship may still be \\ attributed to the original author.\end{tabular}                        & \begin{tabular}[c]{@{}l@{}}Where \textbf{content} of the \\ text matters most, \\such as copyright \& \\ plagiarism of scientific \\articles \end{tabular}                       \\ \midrule

\begin{tabular}[c]{@{}l@{}}The new ship \textit{formed}\\ \textit{with original planks} \\ should be considered \\ original\end{tabular} & \begin{tabular}[c]{@{}l@{}}\textbf{Identity through continuity theory}\\ \citep{wiggins1967identity_through_continuity}: An object \\ retains its identity if there is a \\ continuous chain of physical \\ connections between its various \\ stages. Since the original planks were \\ used to construct the new ship, it \\ creates a direct continuity between \\ the original ship and the new one.\end{tabular} & \begin{tabular}[c]{@{}l@{}}Paraphrasing \textbf{should}\\ \textbf{change} authorship\end{tabular}                                  & \begin{tabular}[c]{@{}l@{}}\textbf{The death of the author theory}\\ \citep{barthes2016death}: Once a text is created, \\ it takes on a life of its own and becomes \\ independent of the author's intention or \\ identity. So, the paraphrasing tool can \\ be considered the author because it is \\ the one actively transforming the \\ original text into a new version.\end{tabular} & \begin{tabular}[c]{@{}l@{}}Where \textbf{style} or \\ probability distribution \\ of text matters most, \\ such as detection of \\AI vs human text\end{tabular} \\ \midrule

\begin{tabular}[c]{@{}l@{}}Both ships exist \\ simultaneously\end{tabular}                                             & \begin{tabular}[c]{@{}l@{}}\textbf{Dual identity theory} \citep{brown2005dual_identity_aquinas}:\\ The ship will be the same in terms of \\ its historical identity (refers to the \\ object's history, narrative, or the \\ sequence of events), while it's physical \\ identity (related to the object's material \\ composition and current state) changes\\ due to the replacement of its planks.\end{tabular}      & \begin{tabular}[c]{@{}l@{}}Authorship as \\ collaborative \\ endeavor as shown in\\ \citep{stillinger1991multiple} \&\\ \citep{chen2011collaborative}\end{tabular} & \begin{tabular}[c]{@{}l@{}}\textbf{Distributed authorship} \citep{ascott2005distributted_authorship}:\\ Authorship is no longer an individual \\ process but is instead shared among \\ multiple entities who contribute to the \\ creation, editing, and dissemination of \\ a text. So, due recognition is extended \\ to both the original author and the \\ paraphraser.\end{tabular}                      & \begin{tabular}[c]{@{}l@{}}If the use of LLM is\\  normalized as writing \\ improvement tool similar \\ to the widespread  \\ integration of grammar \\correction 
tools (\citeauthor{ferris2004grammar}, \\ \citeyear{ferris2004grammar}) \end{tabular}     \\ \bottomrule
\end{tabular}
}
\caption{Different Ship of Theseus ``solutions'' and corresponding authorship scenario in case of paraphrasing}
\label{tab-ship_solution}
\end{table*}
\paragraph{When \textbf{``content''} of text is more important:}
For presentations of original ideas like scientific articles, core \textbf{content} and ideas have the utmost importance. Thus, {\ul LLM paraphrasing should not alter authorship, which should remain with the original content creator, aligning with the \textbf{idea-expression dichotomy}} \citep{samuels1988idea}. For example, \textbf{current ACL policy} mentions that using tools that only assist with language, like Grammarly or spell checkers, \textit{does not need to be disclosed}. However, the stylistic influence of LLMs like ChatGPT could raise flags with AI text detection tools, which authors should consider. 


\paragraph{When \textbf{``style''} of text is more important:}
For AI text detection, authorship centers on ascertaining the source of the text and assessing its conformity to the probability distribution exhibited by the suspected LLM. Therefore, 
expecting detectors to identify heavily paraphrased text as the source poses challenges, as LLM-paraphrased text exhibits the LLM's style. Our alternative ground truth findings further showcase this. Thus, {\ul substantial paraphrasing should change authorship, aligning with the \textbf{death of the author theory}} \citep{barthes2016death}, and {\ul its utility as a perturbation method remains debatable}. However, the ongoing cat-and-mouse game between LLMs and AI text detection necessitates the development of an authorship preservation metric,  a benchmark that any paraphraser should adhere to, and surpassing its threshold would be regarded as a change in authorship.


\paragraph{When both \textbf{``content''} and \textbf{``style''} are important:}
Maintaining the author's unique tone and style is crucial when both the originality of content and creative expression are paramount. The widespread use of ChatGPT or other LLMs in modifying text raises concerns about authenticity, as exemplified by ChatGPT's impact on Clarkesworld, leading to a submission suspension\footnote{\url{https://observer.com/2023/02/science-fiction-magazine-clarkesworld-ban-submission-chatgpt/}}.
\textcolor{black}{
Authors can utilize LLMs for minor proofreading tasks, but it is important to maintain originality and uphold their unique writing style in a substantial portion of the content. Therefore, {\ul assessing authorship should prioritize evaluating the coherent expression of ideas and their flow within the text.}
}

\paragraph{Paraphrasing as AI text generation method?}

Paraphrasing serves as a common technique for data augmentation \citep{BEDDIAR2021100153, okur-etal-2022-data, sharma2022systematic, LI202271, macko2023multitude, cegin2024effects}, particularly valuable in low-resource settings with limited data availability. Our work demonstrates that paraphrasing with an LLM like ChatGPT can align the style with that of the specific LLM. We approximated a \textit{fixed} style for the LLM in a \textit{specific dataset} based on samples generated with minimal prompts. However, this style is modifiable through prompting and varies in other datasets. Therefore, {\ul if LLM paraphrasing is employed for AI text generation, attribution as the author should only occur when a substantial portion of the text is independently generated, not derived from the original prompt}.



\section{Conclusion}
In light of the increasing mainstream popularity of LLMs,
AI text generation and detection have witnessed a surge in research activities. Among the essential components in these domains, paraphrasing holds a significant role. 
This study explores the diverse notions of authorship regarding paraphrasing, 
inspired by the philosophical Ship of Theseus paradox. Our findings suggest that authorship should be task-dependent, and we substantiate our empirical results with theoretical and philosophical perspectives. Given the increasing prevalence of LLMs in generating and enhancing text, our research can provide a sound basis for addressing plagiarism and copyright disputes in the future involving LLMs.

\clearpage
\section*{Limitations}

While our study offers a comprehensive computational and philosophical exploration of  paraphrasing and authorship scenario, it is also important to acknowledge its limitations. When utilizing LLMs as paraphrasers, we employ their default settings with a generic prompt, neglecting specifically tailored instructions. However, in general, the styles of LLM-paraphrased text can vary if instructed to generate in a particular tone or style. 

One limitation of our study is its restriction to the English language. To explore how LLMs and paraphrasing tools in other languages deviate from the source style, further research and expertise in those languages are required. Furthermore, as LLMs can shift the style to conform with the LLM style distribution, it raises a question of whether the reverse is feasible. Can humans paraphrase LLM-generated text to render it with a human-like style? This intriguing avenue warrants additional investigation, and it is critical to include the perspectives of human specialists, including linguists and computational experts, on these ambivalent concepts about authorship present in the current scenario.

\section*{Ethics Statement}
This research was conducted with careful consideration of ethical principles. The tasks of this paper involve paraphrasing existing datasets from humans and LLMs, adhering to their licenses. 

The potential societal impacts of this work, both positive and negative, were contemplated. On the positive side, this research aims to spur thoughtful discussion around emerging issues of authorship attribution and ownership in the age of LLMs. On the negative side, the techniques presented could be misused to misattribute authorship or obfuscate plagiarism intentionally. However, promoting the awareness of these capabilities will enable more informed policy decisions rather than attempts at prohibition, which are unlikely to succeed.

While observational, this study was conducted ethically and does not directly recommend for or against any particular applications of paraphrasing technology. The authors hope that the insights will inform ongoing debates among scholars and policymakers about the proper and fair usage of AI-based writing assistants. Any future research building upon these findings should continue to consider the ethical implications of how text authorship is assigned, quantified, and detected.

\section*{Acknowledgements}

This work was in part supported by U.S. National Science Foundation (NSF) awards \#1820609, \#1950491, \#2131144, and \#2114824, and by \textit{VIGILANT--Vital IntelliGence to Investigate ILlegAl DisiNformaTion}, a project funded by the European Union under the Horizon Europe, GA No. \href{https://doi.org/10.3030/101073921}{101073921}, and \textit{AI-CODE - AI services for COntinuous trust in emerging Digital Environments}, a project funded by the European Union under the Horizon Europe, GA No. \href{https://cordis.europa.eu/project/id/101135437}{101135437}.

Part of the research results was obtained using the computational resources 
provided by the CloudBank (https://www.cloudbank.org/), which was supported by the NSF award \#1925001, and 
those procured in the national project \textit{National competence centre for high performance computing} (project code: 311070AKF2) funded by European Regional Development Fund, EU Structural Funds Informatization of Society, Operational Program Integrated Infrastructure.

\bibliography{main}
\appendix

\clearpage

\section{Methodological Details}
\label{sec_appendix_methods}

This section delves into a nuanced analysis of our methodology, focusing mainly on the datasets and author style models.
\begin{table*}
\renewcommand{\tabcolsep}{2pt}
\centering
\footnotesize
\resizebox{\textwidth}{!}{
\begin{tabular}{@{}|c|c|c|c|@{}}
\toprule
\textbf{Author}                                                 & \textbf{Original ($T^0$)}                                                                                                                                                                                                                                                                                                                                                                                                                                                                                                    & \textbf{ChatGPT paraphrased ($T^1$)}                                                                                                                                                                                                                                                                                                                                                                                                                                                                                                                           & \textbf{ChatGPT paraphrased ($T^2$)}                                                                                                                                                                                                                                                                                                                                                                                                                                                                                                                        \\ \midrule
\textbf{Human}                                                  & \cellcolor[HTML]{9AFF99}\begin{tabular}[c]{@{}l@{}}GANs can generate photo-realistic images \\ from the domain of their training data. \\ However, those wanting to use them for \\ creative purposes often want to generate \\ imagery from a truly novel domain, a task \\ which GANs are inherently unable to do. \\ It is also desirable to have a level of control \\ so that there is a degree of artistic direction \\ rather than purely curation of random results.\end{tabular}                            & \cellcolor[HTML]{9AFF99}\begin{tabular}[c]{@{}l@{}}GANs have the ability to produce realistic \\ images based on the data they were trained \\ on. However, individuals who wish to use \\ GANs for creative purposes often desire to \\ generate images from completely new \\ domains, which GANs are incapable of \\ doing naturally. Additionally, it is ...\\ ----------------------------------------------------\\ S=0.983 = S$^\prime$=0.983 | C=0.972 \textbf{>} C$^\prime$=0.932\end{tabular}                                                                                        & \cellcolor[HTML]{9AFF99}\begin{tabular}[c]{@{}l@{}}GANs can generate realistic images using \\ the provided data, but they cannot naturally \\ create images from different domains. \\ People who want to use GANs for creative \\ purposes often want to generate images \\ from new domains and have some control \\ over the output...\\ ----------------------------------------------------\\ S=0.965 \textbf{<} S$^\prime$=0.973 | C=0.985 \textbf{>} C$^\prime$=0.924\end{tabular}                                                                              \\ \midrule
\textbf{OpenAI}                                                 & \cellcolor[HTML]{FFCCC9}\begin{tabular}[c]{@{}l@{}}\textit{GANs can generate photo-realistic images} \\ \textit{from the domain of their training data.} \\ \textit{However, those wanting to use them for} \\ \textit{creative purposes often want to generate} \\ \textit{imagery from a truly} novel source, without \\ having to manually gather and label \\ training data. In recent years, a technique \\ called StyleGAN has gained popularity \\ as a way to generate novel images....\end{tabular}                                                      & \cellcolor[HTML]{9AFF99}\begin{tabular}[c]{@{}l@{}}GANs have the ability to produce realistic \\ images based on their training data. \\ However, those who desire to use them for \\ creative purposes often seek to generate \\ pictures from a completely new source, \\ without the need to manually collect and \\ categorize data for training....\\ ----------------------------------------------------\\ S=S$^\prime$=0.994 | C=C$^\prime$=0.989\end{tabular}                                                                                                      & \cellcolor[HTML]{9AFF99}\begin{tabular}[c]{@{}l@{}}GANs are capable of generating realistic \\ images based on their training data. However, \\ individuals interested in using GANs for \\ \textit{creative purposes often want to generate} \\ images from a completely new source \\ without the need to manually collect and \\ categorize data for training...\\ -----------------------------------------------------\\ S=S$^\prime$=0.992 | C=C$^\prime$=0.984\end{tabular}                                                                                                \\ \midrule
\textbf{PaLM2}                                                  & \cellcolor[HTML]{9AFF99}\begin{tabular}[c]{@{}l@{}}\textit{GANs can generate photo-realistic images} \\ \textit{from the domain of their training data.} \\ \textit{However, those wanting to use them for} \\ \textit{creative purposes often want to generate} \\ \textit{imagery from a truly} novel domain. One \\ way to achieve this is to use CLIP, a large \\ language model, to provide the text \\ prompt for the GAN....\end{tabular}                                                                                                                  & \cellcolor[HTML]{9AFF99}\begin{tabular}[c]{@{}l@{}}GANs have the ability to produce realistic \\ images based on the data they were trained \\ on. However, individuals seeking to utilize \\ GANs for creative purposes may desire to \\ create images from completely new domains. \\ To accomplish this, one method involves \\ utilizing CLIP, a language model...\\ ----------------------------------------------------\\ S=0.9811 \textbf{<} S$^\prime$=0.9814  | C=0.987 \textbf{>} C$^\prime$=0.885\end{tabular}                                                  & \cellcolor[HTML]{FFCCC9}\begin{tabular}[c]{@{}l@{}}GANs can generate realistic images using \\ trained data, but those who want to use GANs \\ for creative purposes may want to create \\ images from different domains. To do this, \\ one approach is to use CLIP, a large language \\ model, to provide text prompts for the GAN. \\ CLIP can generate ...\\ ----------------------------------------------------\\ S=0.974 \textbf{<} S$^\prime$=0.979 | C=0.964 \textbf{>} C$^\prime$=0.867\end{tabular}                                                          \\ \midrule
\textbf{LLaMA}                                                  & \cellcolor[HTML]{9AFF99}\begin{tabular}[c]{@{}l@{}}\textit{GANs can generate photo-realistic images} \\ \textit{from the domain of their training data.} \\ \textit{However, those wanting to use them for} \\ \textit{creative purposes often want to generate} \\ \textit{imagery from a truly} novel domain, and \\ this is challenging due to the narrow focus \\ of GANs. I will discuss the phenomenon of \\ mode collapse, the difficulty of training a \\ GAN to generate a diverse set of images...\end{tabular}                                         & \cellcolor[HTML]{9AFF99}\begin{tabular}[c]{@{}l@{}}GANs have the ability to produce realistic \\ images that resemble those in their training \\ dataset. However, when using GANs for\\  creative purposes, there is a desire to \\ generate images that belong to a completely \\ new domain. This poses a challenge as \\ GANs tend to have a limited focus. In this \\ discussion, I will explore the concept of ...\\ ----------------------------------------------------\\ S=0.971 \textbf{<} S$^\prime$=0.980 | C=0.977 \textbf{>} C$^\prime$ = 0.882\end{tabular} & \cellcolor[HTML]{FFCCC9}\begin{tabular}[c]{@{}l@{}}GANs possess the capability to create \\ authentic-looking images that resemble those \\ found in their training dataset. However, \\ when GANs are employed for creative \\ purposes, there is a desire to generate images \\ that belong to an entirely new domain. \\ This presents a challenge since GANs tend \\ to have a narrow focus. In this discourse ...\\ ----------------------------------------------------\\ S= 0.966 \textbf{<} S$^\prime$=0.982 | C=0.977 \textbf{>} C$^\prime$=0.880\end{tabular} \\ \midrule
\textbf{Tsinghua}                                               & \cellcolor[HTML]{9AFF99}\begin{tabular}[c]{@{}l@{}}\textit{GANs can generate photo-realistic images} \\ \textit{from the domain of their training data.} \\ \textit{However, those wanting to use them for} \\ \textit{creative purposes often want to generate} \\ \textit{imagery from a truly} novel domain. \\ Unfortunately, it is often difficult to find a \\ suitable domain for training a GAN in this \\ manner. As a result, the image generation \\ quality is often not satisfactory. In this \\ paper, we propose a novel approach ...\end{tabular} & \cellcolor[HTML]{FFCCC9}\begin{tabular}[c]{@{}l@{}}GANs have the ability to create realistic \\ images that belong to the same domain as \\ their training data. However, individuals \\ looking to use GANs for creative purposes \\ often desire to generate images from a \\ completely new domain. Unfortunately, \\ finding a suitable training dataset for this \\ purpose is often challenging, resulting in ... \\ ----------------------------------------------------\\ S=0.898 \textbf{<} S$^\prime$=0.976 | C=0.952 \textbf{>} C$^\prime$ = 0.891\end{tabular} & \cellcolor[HTML]{FFCCC9}\begin{tabular}[c]{@{}l@{}}GANs possess the capability to generate \\ realistic images within the same domain \\ as their training data. However, when it \\ comes to utilizing GANs for creative \\ purposes, individuals often desire the \\ generation of images from an entirely \\ novel domain. Unfortunately, finding an \\ appropriate training dataset for this ...\\ -----------------------------------------------------\\ S=0.884 \textbf{<} S$^\prime$=0.980 | C=0.946 \textbf{>} C$^\prime$ = 0.817\end{tabular}                 \\ \midrule
\textbf{\begin{tabular}[c]{@{}l@{}}Eleuther\\ -AI\end{tabular}} & \cellcolor[HTML]{9AFF99}\begin{tabular}[c]{@{}l@{}}\textit{GANs can generate photo-realistic images} \\ \textit{from the domain of their training data.} \\ \textit{However, those wanting to use them for} \\ \textit{creative purposes often want to generate} \\ \textit{imagery from a truly} novel perspective or \\ with new aesthetic qualities that are not \\ present in existing photos {[}1,2{]}. In this \\ paper we propose an approach which \\ allows us control over "where" input ...\end{tabular}                                               & \cellcolor[HTML]{FFCCC9}\begin{tabular}[c]{@{}l@{}}In this paper, we introduce a method that \\ provides us with the ability to manipulate \\ the input noise in GAN-generated images. \\ This manipulation is achieved through the \\ use of attention maps generated by self-\\ organizing networks (SOM). Our approach \\ allows for the generation of images from  ...\\ -----------------------------------------------------\\ S=0.975 \textbf{<} S$^\prime$=0.978 | C=0.956 \textbf{>} C$^\prime$ = 0.839\end{tabular}                                              & \cellcolor[HTML]{FFCCC9}\begin{tabular}[c]{@{}l@{}}This paper presents a technique that enables \\ us to control the input noise in images \\ generated by GANs. By utilizing attention \\ maps created by self-organizing networks \\ (SOM), we are able to manipulate the noise \\ and generate images with distinct \\ viewpoints and artistic qualities, surpassing \\ ----------------------------------------------------\\ S=0.974 \textbf{<} S$^\prime$=0.981 | C=0.952 \textbf{>} C$^\prime$ = 0.829\end{tabular}                                              \\ \midrule
\textbf{\begin{tabular}[c]{@{}l@{}}Big-\\ Science\end{tabular}} & \cellcolor[HTML]{9AFF99}\begin{tabular}[c]{@{}l@{}}\textit{GANs can generate photo-realistic images} \\ \textit{from the domain of their training data.} \\ \textit{However, those wanting to use them for} \\ \textit{creative purposes often want to generate} \\ \textit{imagery from a truly} novel viewpoint. \\ Our paper describes an approach based \\ on multi-view learning that enables one-\\ to-many style transfer when generating \\ artistic photographs using untrained DNN ...\end{tabular}                                                     & \cellcolor[HTML]{FFCCC9}\begin{tabular}[c]{@{}l@{}}GANs have the ability to generate realistic \\ images based on the training data they \\ receive. However, for creative purposes, \\ it is often desirable to generate images \\ from unique viewpoints. Our research \\ paper presents a method that utilizes multi-\\ view learning to enable one-to-many style ... \\ -----------------------------------------------------\\ S=0.981 = S$^\prime$=0.981 | C=0.981 \textbf{>} C$^\prime$ = 0.901\end{tabular}                                                       & \cellcolor[HTML]{FFCCC9}\begin{tabular}[c]{@{}l@{}}GANs possess the capability to generate \\ lifelike images based on the training data \\ they receive. However, when it comes to \\ artistic purposes, there is often a desire to \\ produce images from unique perspectives. \\ Our research paper introduces a method \\ that utilizes multi-view learning to enable ...\\ ----------------------------------------------------\\ S=0.980 \textbf{<} S$^\prime$=0.981 | C=0.971 \textbf{>} C$^\prime$ =0.875\end{tabular}                                          \\ \bottomrule
\end{tabular}
}
\caption{Example of our paraphrased corpus (\textbf{sci\_gen} dataset). The original text ($T^0$) from each author was paraphrased subsequently by \textbf{ChatGPT} to generate $T^1$, $T^2$, ... The \textit{italic part} of $T^0$ was the \textbf{prompt} for generating from other authors (LLMs). \textbf{S} and \textbf{S$^\prime$} identify the style similarity with the original text version ($T^0$) from the corresponding author and \textbf{ChatGPT}, respectively. Similarly, \textbf{C} and \textbf{C$^\prime$} show the content similarity. \textcolor[HTML]{9AFF99}{Green cells} identify that it was correctly predicted as the respective author (Finetuned BERT), whereas \textcolor[HTML]{FFCCC9}{red} shows mis-classifications. We observe \textbf{C>C$^\prime$}, whereas \textbf{S} is mostly less than \textbf{S$^\prime$} and decreases from the previous iteration.}
\label{tab_dataset_example}
\end{table*}

\subsection{Dataset examples}
For a clearer understanding of multiple authors, paraphrasers, and iterations, Table \ref{tab_dataset_example} provides an example from our paraphrased corpus. As discussed in Section \ref{sec_method}, we use identical samples from all authors as training samples to ensure fair training and style model creation, mitigating any potential bias from the content of the texts and making the classification task more challenging and realistic in settings where multiple authors have writings on the same topic.

\begin{figure*}[h]
    \centering
    \includegraphics[width=\linewidth]{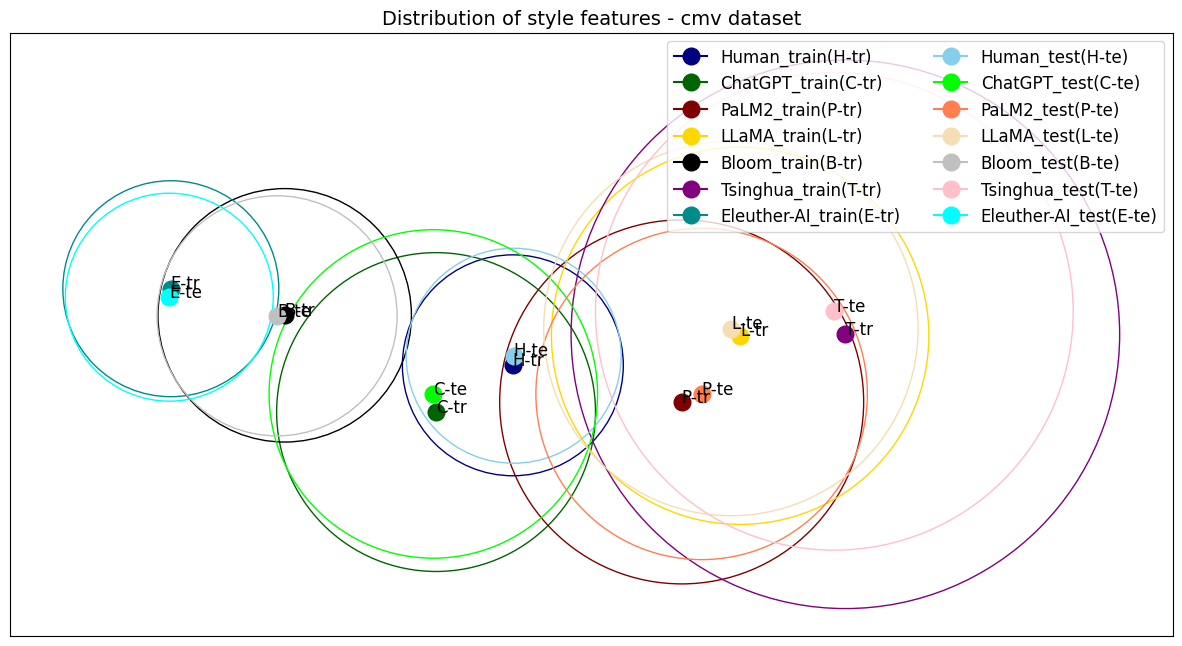}
    \caption{PCA visualization of the style features for train samples and original ($T^0$) test samples for different authors. The point represents the mean of the distribution, and the circle approximates the distribution (containing 90\% of all samples). }
    \label{fig_train_test_distributions}
\end{figure*}

While primarily used for author attribution tasks to validate both traditional and alternative perspectives of ground truths, we also leverage a subset of our datasets to address the human vs. AI text detection problem as follows.

\begin{itemize}
    \item \textbf{Normal:} In a normal scenario, without paraphrasing, we designate $T^0$(human) as \textbf{human} and $T^0$(ChatGPT) as \textbf{AI} text for each dataset.
    \item \textbf{Traditional:} In the traditional setting, where paraphrasing maintains authorship, we designate ChatGPT paraphrased (after the first iteration) of the original human text, $T^1$(human) as \textbf{human} text, and similarly $T^1$(ChatGPT) as \textbf{AI} text.
    \item \textbf{Alternative:} In the alternative scenario, where paraphrasing alters authorship, we label ChatGPT paraphrased (after the first iteration) of the original human text,  $T^1$(human), as \textbf{AI} text, while the original version of the human text,  $T^0$(human), is designated as \textbf{human} text.
\end{itemize}

\subsection{Validity of author style model}
\label{subsec_style_model_validity}
Section \ref{sec_method} explains our motivation for opting for feature-based methods to approximate a style model for any author. We substantiate our choice both statistically  and based on classification performance. It is essential to note that our style model is approximated individually for each dataset. Therefore, the human style model in xsum, for instance, differs from the human style model in the cmv dataset. Figure \ref{fig_train_test_distributions} illustrates that the distribution of train and test samples for any specific author appears similar in the 2D space, validating that the style model from $T^0$ should approximate the style of that particular author.

\begin{figure*}[!htb]
    \centering

  \centering
  \begin{subfigure}{0.49\textwidth}
    \centering
    \includegraphics[width=\textwidth]{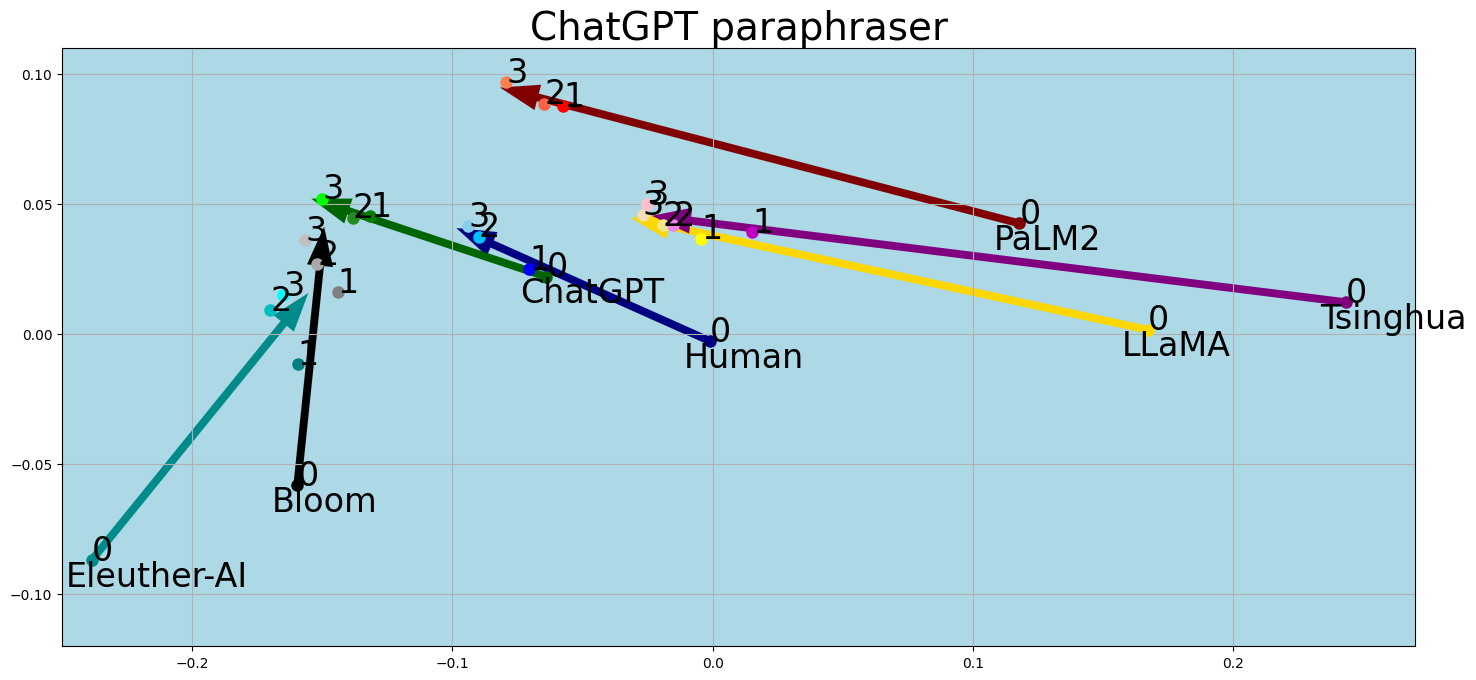}
  \end{subfigure}
  \hfill
  \begin{subfigure}{0.49\textwidth}
    \centering
    \includegraphics[width=\textwidth]{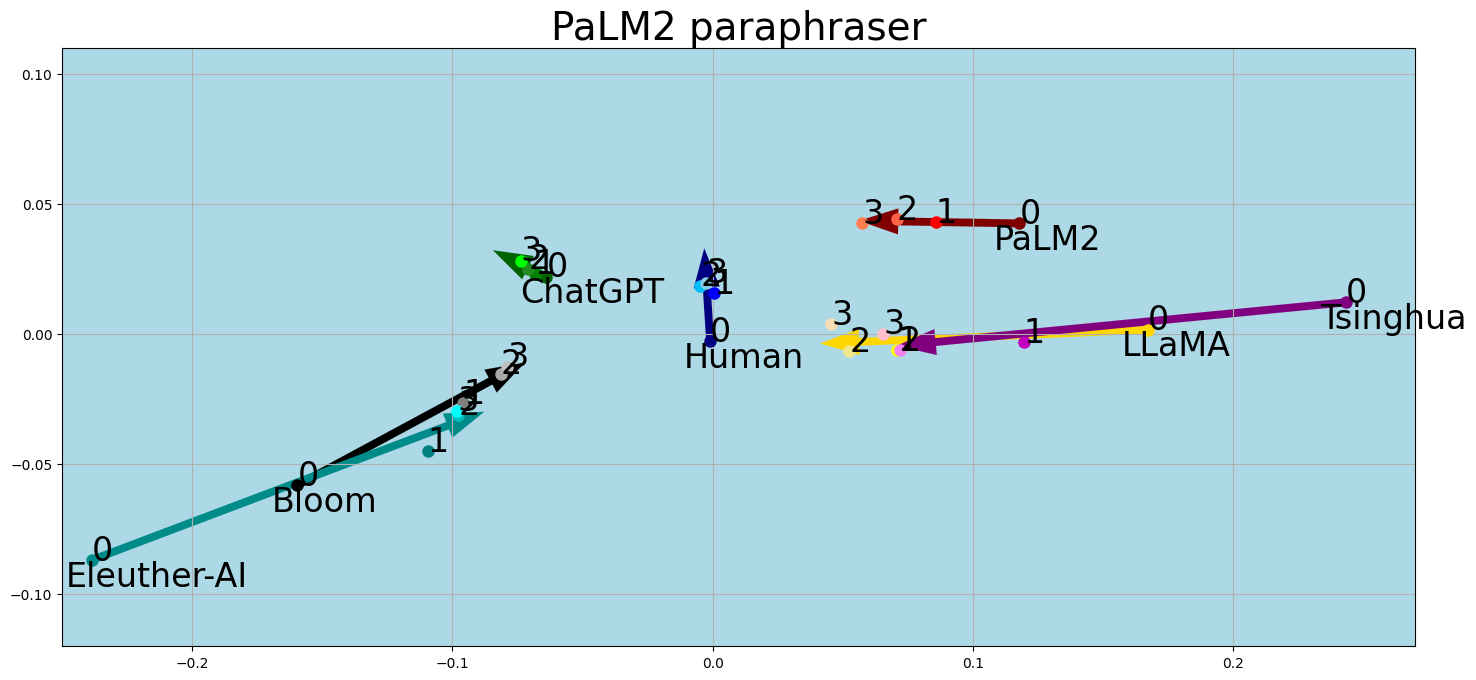}
  \end{subfigure}
  
  
  
  \begin{subfigure}{0.49\textwidth}
    \centering1
    \includegraphics[width=\textwidth]{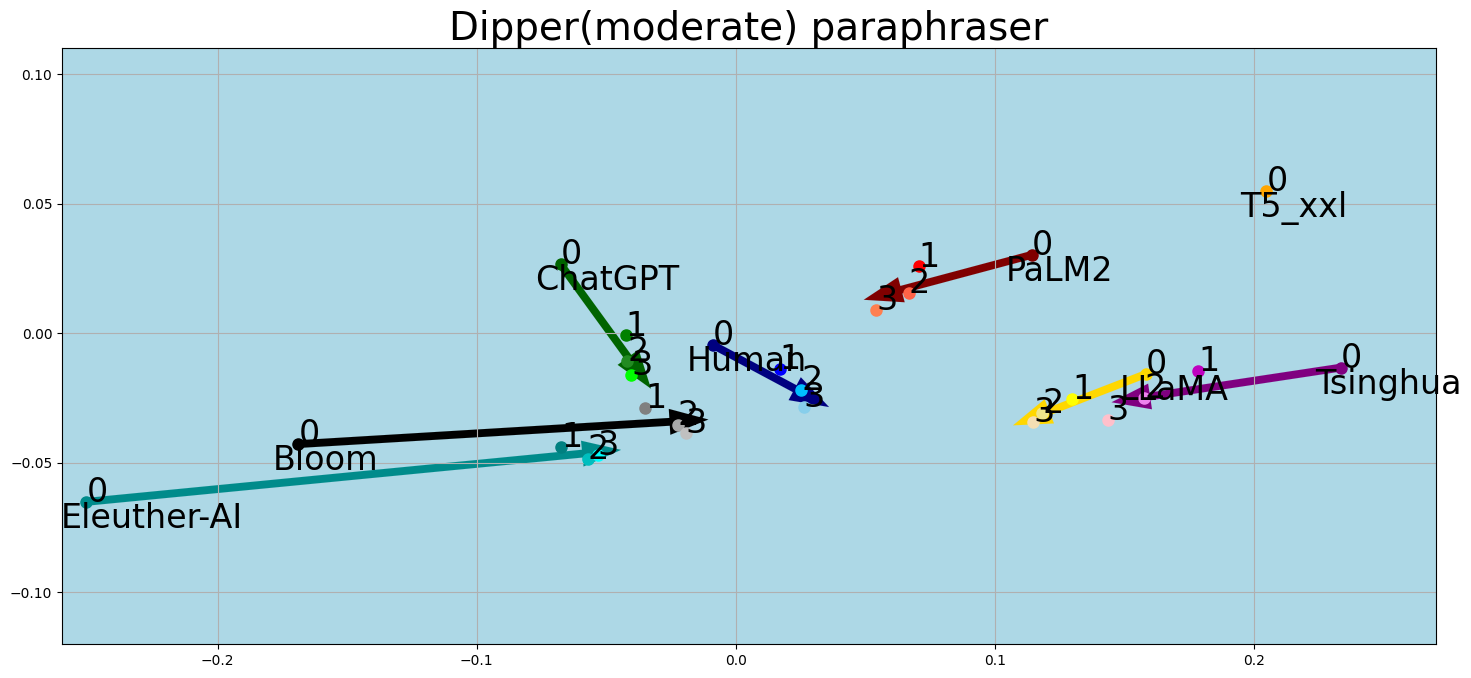}
  \end{subfigure}
  \hfill
  \begin{subfigure}{0.49\textwidth}
    \centering
    \includegraphics[width=\textwidth]{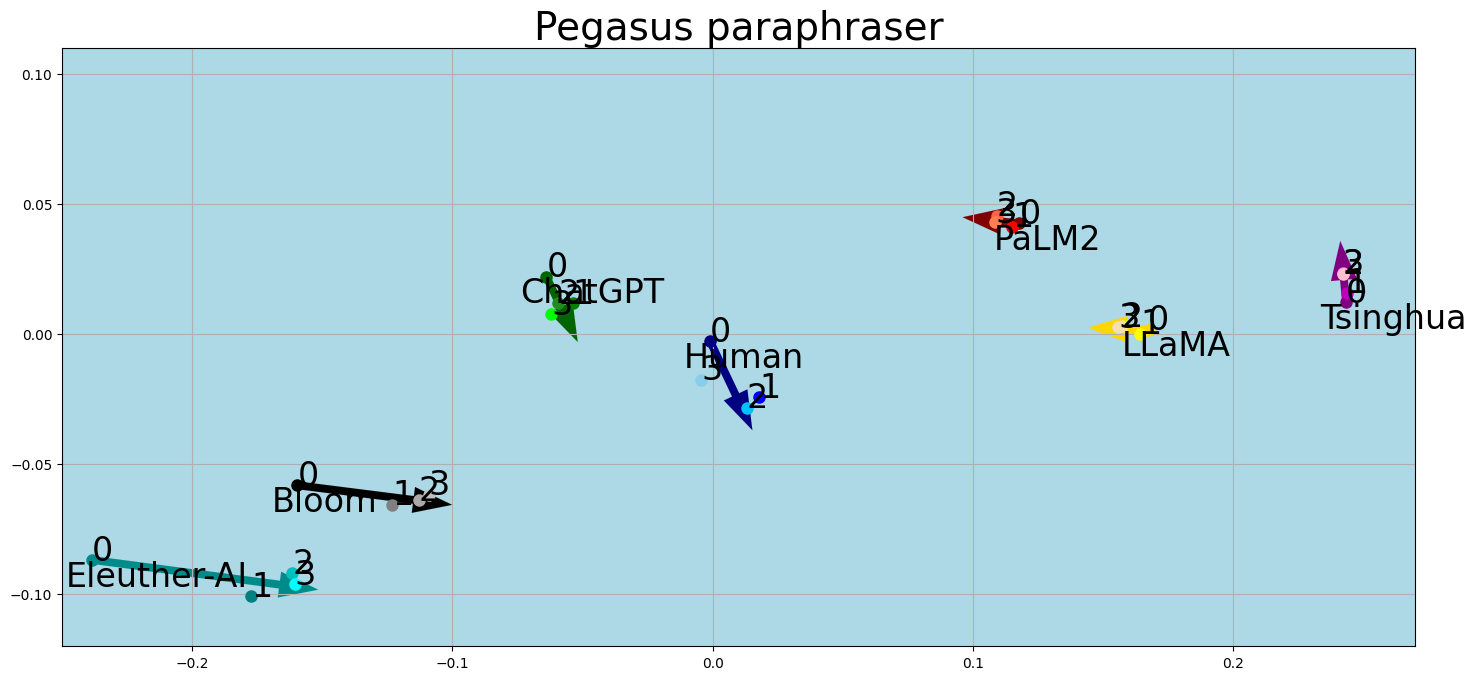}
  \end{subfigure}

\caption{PCA visualization of author style shifts in \textbf{cmv} dataset using various paraphrasers. Points represent the center of samples for each author and version (0-original, 1,2,3-subsequent paraphrased). Arrows indicate style shifts. ChatGPT significantly alters the style of all authors, centralizing them around ChatGPT's style; PaLM2 exhibits a similar though less pronounced behavior. Pegasus induces minimal changes, while Dipper, despite a substantial shift, diverges from the style of its training LM, T5-xxl. However, we do not observe any content shift from the original texts for all paraphrasers, as depicted in Figure \ref{fig_content_style_shift} for ChatGPT. }
\label{fig_pca_visualization}
\end{figure*}

\begin{table*}[h]
\centering
\renewcommand{\tabcolsep}{2pt}
\footnotesize
\resizebox{\textwidth}{!}{
\begin{tabular}{@{}l|lllllll@{}}
\toprule
\textbf{Method}           & \textbf{xsum} & \textbf{tldr} & \textbf{sci\_gen} & \textbf{cmv} & \textbf{wp} & \textbf{eli5} & \textbf{yelp} \\ \midrule
\textbf{BERT(best model)} & 0.72          & 0.74          & 0.77              & 0.82         & 0.81        & 0.78          & 0.79          \\
\textbf{Style model}      & 0.67          & 0.63          & 0.7               & 0.8          & 0.76        & 0.71          & 0.72          \\
\textbf{WritePrints only}  &  0.64 \dec{4.5} & 0.61 \dec{3.2} & 0.68 \dec{2.9} & 0.78 \dec{2.5} & 0.75 \dec{1.3} & 0.7 \dec{1.4} & 0.7 \dec{2.8}          \\
\textbf{LIWC only}        &  0.52 \dec{22.4} & 0.48 \dec{23.8} & 0.57 \dec{18.6} & 0.67 \dec{16.2} & 0.64 \dec{15.8} & 0.56 \dec{21.1} & 0.57 \dec{20.8}          \\ \bottomrule
\end{tabular}
}
\caption{Ablation study for performance of style model in authorship attribution. \textcolor{red}{$\downarrow$} denotes performance drop  if a specific component is removed, compared to style model.  }
\label{tab_style_model_ablation}
\end{table*}

We also employ the style model as a stylometry measure for author attribution. Despite its simplicity, it achieves the second-best performance in the original version ($T^0$), with a slight decrease compared to the best-performing Fine-tuned BERT. The ablation study (Section \ref{sec_appendix_ablation}) demonstrates that utilizing LIWC and WritePrint yields better results than considering them individually. Future work will focus on identifying feature importance to have a more nuanced understanding of style and how paraphrasing impacts it.

\section{Experimental Details}
\label{sec_experimental_details}

This section details our authorship attribution and AI text detection methods, encompassing a pre-processing step for compatibility. We eliminated samples falling below-specified thresholds (100 word counts for authorship attribution and 200 for AI text detection). To mitigate randomness, we also conducted the experiments five times for each text classification task, reporting the average in all tables. Figures \ref{fig_pca_visualization}  shows how the style and content shift for different paraphrasers.

\subsection{Authorship attribution methods}
Since our authorship attribution is a seven-class text classification problem, we rely on the supervised/finetuned method for classification. 
\paragraph{Finetuned BERT:} As fine-tuned language models have been state of the art in text classification tasks, we fine-tune BERT (\textit{bert-base-cased}) on each dataset training set and evaluate it on the test set.

\paragraph{Stylometry:} We employ our style model that combines LIWC \citep{pennebaker2001LIWC} and WritePrint \citep{abbasi2008writeprints_DT} features with an XGBoost classifier as the stylometry method. LIWC analyzes text using over 60 categories representing a range of social, cognitive, and affective processes. WritePrint extracts lexical and syntactic features, encompassing char, word, letter, bigram, trigram, vocabulary richness, pos-tags, punctuation, and function words. In total, we utilize 623 features. 

\paragraph{GPT-who:} GPT-who \citep{venkatraman2023gptwho}, a psycho-linguistically-aware multi-class domain-agnostic statistical-based detector, utilizes UID-based features to capture a unique statistical signature. Initially designed for AI text detection, we repurpose it for our multi-class settings. The UID features are generated through inference from GPT-2, and an Logistic Regression (LR) model is trained on the dataset.

\paragraph{TF-IDF:} 
We employ character n-grams (n=2 to 5) represented by TF-IDF scores in conjunction with an LR classifier. While n-grams excel in traditional authorship attribution tasks \citep{Tyo_Dhingra_Lipton_2022_AV, tripto2023hansen}, their performance is comparatively lower in our dataset since all authors have training samples on similar topics.

\begin{figure*}[h]
  \begin{subfigure}{0.32\textwidth}
    \centering
    \includegraphics[width=\linewidth]{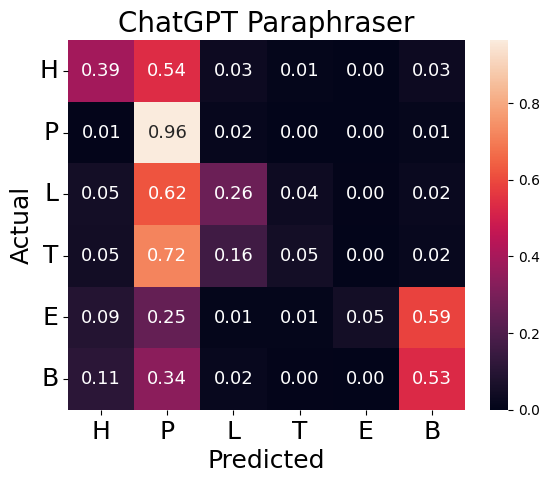}
  \end{subfigure}
    \hfill
  \begin{subfigure}{0.32\textwidth}
    \centering
    \includegraphics[width=\linewidth]{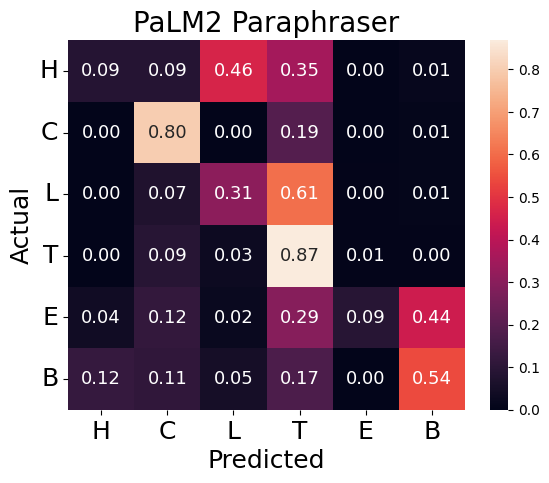}
  \end{subfigure}
    \hfill
  \begin{subfigure}{0.32\textwidth}
    \centering
    \includegraphics[width=\linewidth]{figures/cm_dipper.png}
  \end{subfigure}
  \caption{Confusion matrix for the Fine-tuned BERT classifier when the respective paraphraser LLM (ChatGPT or PaLM2) is left out from the training. (H: Human, C: ChatGPT, P: PaLM2, L: LLAMA, T: Tsinghua, E: Eleuther-AI, B: Bloom). Even in this scenario, misclassifications are aligned to another LLM (although different from the paraphrasing LLM) compared to the Dipper paraphraser.}
  \label{fig-confusion-matrix-ablation}
\end{figure*}

\subsection{AI text detection methods:}
While fine-tuning language models enhances AI text detection performance on specific datasets \citep{HSCBZ23mgtbench}, depending solely on this approach is not a comprehensive solution, given the rapid growth of LLMs and their generated texts. Therefore, we restrict ourselves to one fine-tuned method and incorporate mostly zero-shot/statistical detectors in our experiments.

\paragraph{Finetuned BERT:} Like authorship attribution, we finetune our BERT model for two classes (human and AI) and evaluate performance on the test sets. 

\paragraph{DetectGPT:} DetectGPT \citep{mitchell2023detectgpt} is a zero-shot AI text classifier that generates perturbed samples from the original text and calculates their probabilities under the model parameters. We utilize T5-3b as the mask-filling model and generate 50 samples as perturbed examples.

\paragraph{GPT-Zero} GPT-Zero \citep{GPTzero} employs perplexity to gauge the complexity of the text and Burstiness to assess variations in sentences, determining whether the text is AI-generated.

\paragraph{RoBERTa-large:} Initially developed as the GPT-2 output detector, this model was created through fine-tuning a RoBERTa large model using the outputs of the 1.5B-parameter GPT-2 model \citep{conneau2019xlm-roberta}.

\paragraph{LongFormer:} Longformer \citep{Beltagy2020Longformer}, a modified Transformer architecture, overcomes the limitations of traditional transformer models by efficiently handling more than 512 tokens. It employs an attention pattern scaling linearly with sequence length, facilitating the processing of longer documents. The Longformer used in our study \citep{li2023deepfake} is based on a comprehensive dataset comprising 447,674 human-written and machine-generated texts.

\begin{table*}[h]
\centering
\renewcommand{\tabcolsep}{2pt}
\footnotesize
\resizebox{0.75\textwidth}{!}{
\begin{tabular}{@{}l|l|l@{}}
\toprule
\textbf{Dataset}                   & \textbf{Paraphraser} & \textbf{Text $T^0 \Rightarrow T^1 \Rightarrow T^2 \Rightarrow T^3 \Rightarrow T^4 \Rightarrow T^5 \Rightarrow T^6  \Rightarrow T^7$}                      \\ \midrule
\multirow{2}{*}{\textbf{xsum}}     & \textbf{ChatGPT}     & 0.72 $\Rightarrow$ 0.26 $\Rightarrow$ 0.24 $\Rightarrow$ 0.22 $\Rightarrow$ 0.21 $\Rightarrow$ 0.2 $\Rightarrow$ 0.21 $\Rightarrow$ 0.18  \\
                                   & \textbf{Dipper}      & 0.7 $\Rightarrow$ 0.27 $\Rightarrow$ 0.27 $\Rightarrow$ 0.24 $\Rightarrow$ 0.22 $\Rightarrow$ 0.19 $\Rightarrow$ 0.2 $\Rightarrow$ 0.16   \\ \midrule
\multirow{2}{*}{\textbf{cmv}}      & \textbf{ChatGPT}     & 0.79 $\Rightarrow$ 0.3 $\Rightarrow$ 0.24 $\Rightarrow$ 0.24 $\Rightarrow$ 0.24 $\Rightarrow$ 0.23 $\Rightarrow$ 0.22 $\Rightarrow$ 0.22  \\
                                   & \textbf{Dipper}      & 0.76 $\Rightarrow$ 0.45 $\Rightarrow$ 0.41 $\Rightarrow$ 0.36 $\Rightarrow$ 0.32 $\Rightarrow$ 0.28 $\Rightarrow$ 0.27 $\Rightarrow$ 0.27 \\ \midrule
\multirow{2}{*}{\textbf{sci\_gen}} & \textbf{ChatGPT}     & 0.74 $\Rightarrow$ 0.39 $\Rightarrow$ 0.33 $\Rightarrow$ 0.3 $\Rightarrow$ 0.31 $\Rightarrow$ 0.29 $\Rightarrow$ 0.31 $\Rightarrow$ 0.32  \\
                                   & \textbf{Dipper}      & 0.73 $\Rightarrow$ 0.37 $\Rightarrow$ 0.22 $\Rightarrow$ 0.16 $\Rightarrow$ 0.18 $\Rightarrow$ 0.16 $\Rightarrow$ 0.11 $\Rightarrow$ 0.15 \\ \bottomrule
\end{tabular}
}
\caption{Performance of Finetuned BERT classifier up to seven paraphrasing iterations (traditional perspective).}
\label{tab_multiple_paraphrasing}
\end{table*}

\section{Ablation Study}
\label{sec_appendix_ablation}

For ablation study, we have conducted several experiments supporting our decisions and/or findings.

\paragraph{Paraphrasing more than three times}
While the Ship of Theseus undergoes numerous modifications before posing the paradox, our experimental constraints lead us to limit paraphrasing iterations to three for most findings. Beyond the initial iterations, we observe minimal shifts in performance and style/content changes. To further investigate, we conducted an ablation study by paraphrasing a subset of our datasets up to seven times. Table~\ref{tab_multiple_paraphrasing} presents the performance of the Finetuned BERT classifier (best model) after seven paraphrasing iterations by two paraphrasers (ChatGPT as an LLM paraphraser and Dipper as a PM paraphraser). The results support our choice of three iterations in experiments as sufficient, as the drop in classification performance becomes negligible for the later versions. Notably, Dipper's paraphrasing leads to a more rapid performance decline than ChatGPT.

\paragraph{Style model without all components}
In Table~\ref{tab_style_model_ablation}, a comparison of using different style models for authorship attribution is provided. It shows that the used combined style model is more suitable than using just the existing WritePrints or LIWC features.

\paragraph{Misclassifications when paraphrasing LLM is not an author}
While Dipper paraphraser causes slightly more performance drops and style shifts compared to ChatGPT and PaLM2 paraphrasers, its misclassifications exhibit a more uniform distribution across all classes (Figure~\ref{fig-confusion-matrix}) , in contrast to LLM paraphrasers. This phenomenon may be attributed to the absence of a PM-specific class label. To address this issue, we excluded ChatGPT/PaLM2 from classifier training and examined the distribution of ChatGPT/PaLM2-generated texts among other classes after classification. Figure~\ref{fig-confusion-matrix-ablation} presents such authorship attribution results in the form of a confusion matrix (in comparison to Figure~\ref{fig-confusion-matrix}). Despite this exclusion, LLM paraphrasers still converge the style to a specific LLM, albeit different from the paraphrasing LLM, as it is excluded from the possible authors.

\end{document}